\documentclass{article}

\PassOptionsToPackage{numbers, compress}{natbib}
\usepackage[eandd, final]{neurips_2026}
\usepackage[utf8]{inputenc} % allow utf-8 input
\usepackage[T1]{fontenc}    % use 8-bit T1 fonts
\usepackage{hyperref}       % hyperlinks
\usepackage{url}            % simple URL typesetting
\usepackage{booktabs}       % professional-quality tables
\usepackage{amsfonts}       % blackboard math symbols
\usepackage{nicefrac}       % compact symbols for 1/2, etc.
\usepackage{microtype}      % microtypography
\usepackage{xcolor}         % colors

\usepackage{graphicx}                       % graphics n stuff
\usepackage{amsmath}                        % math n stuff
\usepackage{pifont}
\usepackage{caption}
\usepackage{tabularray}
\usepackage{tabularx}
\usepackage{algorithm}
\usepackage{algorithmic}
\usepackage{soul}
\usepackage{bm}
\usepackage{multirow}
\usepackage{xspace}
\usepackage{subcaption}
\usepackage[capitalize,noabbrev]{cleveref}
\usepackage[dvipsnames, table]{xcolor}
\newcommand{\cmark}{$\checkmark$}
\newcommand{\xmark}{$\times$}

\newcommand{\modelname}{H2M2\xspace}    % MetaHuman Hair Matting Model
\newcommand{\datasetname}{SynthHair\xspace}   % MetaHuman Hair Dataset

\title{SynthHair: Leveraging MetaHumans for a High-Quality 4K Hair Matting Dataset}

\author{
  \textmd{
    \large Markus Karmann\textsuperscript{1} \qquad
    Shile Li\textsuperscript{2} \qquad
    Qi Zhang\textsuperscript{3} \qquad
    Peng-Tao Jiang\textsuperscript{3} \qquad
  } \\[0.5ex]
  \textmd{
    \large Bo Li\textsuperscript{3} \qquad
    Puneet K. Dokania\textsuperscript{1}\qquad
    Philip Torr\textsuperscript{1} \qquad
    Onay Urfalioglu\textsuperscript{2}
  } \\[3ex]
  \textsuperscript{1}University of Oxford \qquad
  \textsuperscript{2}vivo Tech Research GmbH \\[0.5ex]
  \textsuperscript{3}vivo BlueImage Lab, vivo Mobile Communication Co., Ltd., China 
}

\begin{document}

\maketitle

\begin{abstract}
  Recent progress in computer vision enables dense alpha matte predictions at resolutions of 4K and higher.
While portrait matting has benefited from large-scale, high-resolution datasets, progress in hair matting remains limited due to the difficulty of annotating sub-pixel fine hair details.
In this work, we analyze critical limitations in image resolution and annotation quality of such existing datasets and introduce our own synthetic hair matting dataset, \textbf{SynthHair}, which consists of 16k RGBA portrait images in $4096 \times 4096$ image resolution, containing labels for hair matting, as well as hair categorization, growth direction, and depth estimation.
We create this dataset using MetaHuman Creator and Unreal Engine to obtain highly detailed and realistic renders of human faces with complex hair structures.
We adapt a generalized version of the isoperimetric inequality quotient (IPQ) that we call the \textbf{soft isoperimetric inequality quotient (SIPQ)}, as a scale-invariant label complexity metric that enables comparison of both segmentation and matte labels.
Using this metric, we demonstrate that hair matting complexity increases significantly with resolution, reaching an SIPQ of 734.75 at 4K, in contrast with portrait matting datasets at similar resolutions, which reach an SIPQ of ${\sim}$10.
To validate our data, we train separate matting models on SynthHair and existing real-world datasets.
Our results demonstrate strong domain generalization and show clear improvements in high-resolution images.
A/B testing further confirms this, as our mattes are preferred in at least 85\% of comparisons on high-resolution images.
Code and data: \url{https://github.com/mkarmann/synth-hair}
\end{abstract}
\begin{figure}[h!]
  \centering{\includegraphics[width=\textwidth]{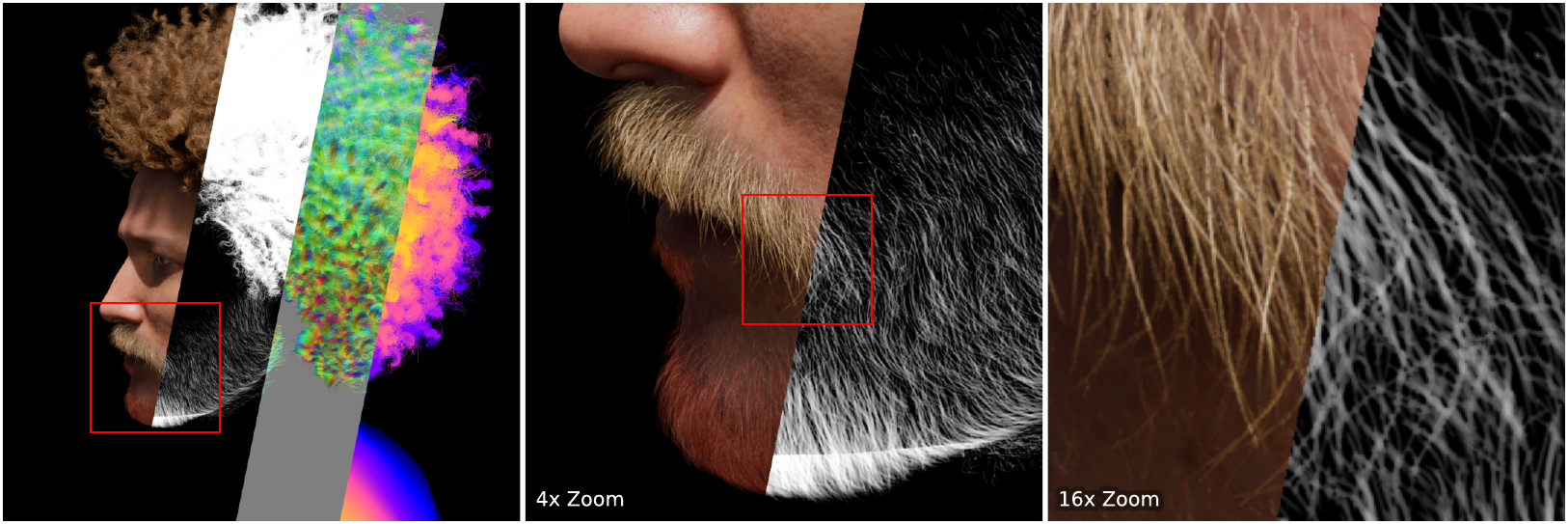}}
  \caption{Our novel \datasetname dataset provides 16k RGBA images with hair matting, category, growth direction, and depth labels. Close-ups show the high level of detail of images and annotations.}
  \label{fig:thumbnail}
\end{figure}

\section{Introduction}
\label{sec:introduction}
Accurate hair matting remains one of the most visually demanding problems in computer vision.
Recent advances in image and video processing enable segmentation and alpha matting of very fine details at resolutions such as 4K and beyond~\cite{SparseMat, RobustVideoMatting}.
While advances in high-resolution portrait matting and image segmentation have been driven by high-quality datasets~\cite{DAViD, ThinObject5k, MODNet, SparseMat}, progress in hair segmentation and matting remains limited by the lack of such datasets.

The human head contains around 100k individual hairs~\cite{HairModelingSurvey}, many being thinner than a single pixel even at ultra-high resolutions.
It is, therefore, practically infeasible to create precise labels manually.
Capture setups like green screens can not be utilized to generate ground truth data, as a contrasting screen cannot be placed between facial hair and the human face.
Existing workflows often incorporate tools such as Photoshop~\cite{Photoshop} filters to aid the annotation process.
These introduce artifacts in the presence of gray hair, complex backgrounds, or general problems in cases where the hair color is not uniform or clearly distinguishable.
Furthermore, separating hair from skin is particularly challenging due to shadows and the presence of individual, non-overlapping hair strands, especially in short haircuts and beards.
In this work, we analyze the impact of these challenges on existing hair segmentation and matting datasets.
To do this, we adapt a generalized version of the isoperimetric inequality quotient, called \textbf{SIPQ}, to show the significant increase in complexity at higher resolutions.

To enable further advancement in ultra-high resolution hair matting, we generate a novel synthetic dataset named \datasetname with the Unreal Engine~5.6~\cite{UnrealEngine} and the integrated MetaHuman Creator 1.1.0~\cite{MetaHumanCreator}.
The dataset consists of rendered images with $4096 \times 4096$ resolution and highly accurate labels (see \cref{fig:thumbnail}).
Existing datasets typically provide binary segmentation masks~\cite{KHairstyle, Figaro1k, FaceSynthetics, CelebAMask-HQ} as ground truth annotation.
These fail to capture very thin and translucent hair strands; we therefore provide \textbf{continuous alpha mattes} as labels instead.
To the best of our knowledge, our dataset is also the first to provide comprehensive annotations for all major facial hair regions, including scalp hair, eyebrows, eyelashes, mustaches, and beards.
In addition to hair matting labels, we further provide labels for \textbf{hair category classification}, \textbf{3D growth direction}, and \textbf{depth estimation}, as these provide valuable cues for tasks such as 3D hair reconstruction and strand geometry recovery.
Since this work primarily focuses on hair matting, we provide initial results using these additional labels in the appendix.
In general, our main contributions are the following:
\begin{itemize}
    \item \textbf{\datasetname}: A novel dataset consisting of 16,000 RGBA images with hair alpha matte, category, growth direction, and depth labels in ultra-high $4096 \times 4096$ image resolution.
    \item We adapt the soft isoperimetric inequality quotient (SIPQ) to compare the structural complexity of alpha mattes and binary segmentation masks.
    \item \textbf{Dataset Validation}: We demonstrate the practical utility of SynthHair by training a baseline model (\modelname), showing that our synthetic labels enable strong domain generalization to real-world images and better hair matting in 4K image resolution.
\end{itemize}

\section{Related Work} \label{sec:related_work}
\paragraph{Hair matting and segmentation}
Obtaining high-quality datasets for hair matting and segmentation is challenging due to the fine structure of hair, resulting in a time consuming and error-prone annotation process.
HairMatteNet~\cite{HairMatteNet} uses crowd-sourced annotations, which are often imprecise and inconsistent.
Figaro-1k~\cite{Figaro1k} provides high-quality manually annotated masks for hair segmentation for a small set of 1050 images with a limited resolution of about $600 \times 700$ pixels.
The K-Hairstyle dataset~\cite{KHairstyle} consists of a large quantity of 500k images, but provides only polygonal hair segmentation masks, annotated at a resolution of $512 \times 512$.
With an image dimension of $512 \times 512$, the strand annotation dataset HiSa~\cite{HairStep} uses an unreleased matting tool to obtain their ground truth hair matting labels for a total of 1250 images.
Hair segmentation is often a part of the more general face attribute segmentation task.
The CelebAMask-HQ dataset~\cite{CelebAMask-HQ} contains 30k $1024 \times 1024$ annotated images with segmentation labels at a resolution of $512 \times 512$.
Fully synthetic datasets such as Face Synthetics~\cite{FaceSynthetics} advance the label quality by rendering pixel-perfect hair masks through photorealistic ray tracing~\cite{FaceSynthetics, DAViD}.
The dataset consists of 100k samples, all generated at a dimension of $512 \times 512$.
Correctly identifying hair is also a significant challenge in portrait matting \cite{SparseMat, MODNet, GenerativeVideoMatting}.
Generative Video Matting~\cite{GenerativeVideoMatting} introduces the synthetic SynHairMan dataset to pre-train their model on portrait alpha mattes with accurate hair annotation, emphasizing the importance of precise hair labels.
Building on this trend, modern graphics engines such as Unreal Engine~\cite{UnrealEngine} have emerged as powerful tools for generating large-scale, photorealistic datasets with perfect ground truth labels.
Unreal Engine provides realistic real-time visuals for tasks such as autonomous driving \cite{CARLA, AirSim} and embodied systems \cite{UnrealZoo}.
It allows generating datasets with perfect ground truth labels for tasks estimating depth~\cite{UnrealCV, UnrealStereo4K}, surface normals \cite{UnrealCV}, bonding-boxes~\cite{Boundless}, and segmentation masks~\cite{UnrealCV, FallingThings}.
Leveraging these capabilities, we use Unreal Engine in combination with the MetaHuman Creator~\cite{MetaHumanCreator} to generate a new 4K-resolution hair matting dataset.

\paragraph{Measuring Label Complexity}
For binary segmentation, \cite{DIS5k} uses three metrics based on polygonal reconstruction of the segmentation boundary.
The number of dominant points, which is the number of polygon vertices, the number of disconnected Contours, and the IPQ.
The IPQ, defined as in \cite{DIS5k}, has a value range of $[1, \infty)$, with $1$ being a circle.
The work \cite{SPIN} uses the reciprocal of this IPQ to measure the boundary complexity of part segmentation labels.

\paragraph{Training on synthetic data}
Depth Anything v2~\cite{DepthAnythingV2} uses no real labeled data in its training process. Instead, the pre-trained DINOv2~\cite{DINOv2} is used as a backbone to train a teacher model on synthetic data.
The teacher model is then used to create pseudo-labels for real images in order to train a student model.
Recent works such as DAViD~\cite{DAViD} show promising results without utilizing a pre-trained model and training only on synthetic data.
For our experiments, we utilize the strong generalization capabilities of DINOv2 and rely on the architecture of DAViD.

\paragraph{Orientation and Depth Map Labels}
In 3D hair reconstruction, methods often utilize intermediate representations such as orientation- or depth maps ~\cite{HairStep, StrandAccurateMultiViewHairCapture, CaptureOfHairGeometryFromMultipleImages, HairPhotobooth, RobustHairCaptureUsingSimulatedExamples, MonoHair} to aid the 3D hair strand estimation process.
Methods such as HairStep~\cite{HairStep} use a neural network to predict a 2D orientation map from a single image.
In order to train such a model, the authors created a custom annotation tool to manually annotate real photographs, which takes on average 1 hour per image.
In earlier works, depth maps are obtained by triangulation of multiple views~\cite{StrandAccurateMultiViewHairCapture, CaptureOfHairGeometryFromMultipleImages}, while newer works obtain depth maps from single images by using neural networks~\cite{HairStep, DepthAnythingV1, DepthAnythingV2}.
We therefore also provide labels for both the 3D hair growth direction and depth estimation to aid training such models.
\begin{figure*}
  \centering{\includegraphics[width=\textwidth]{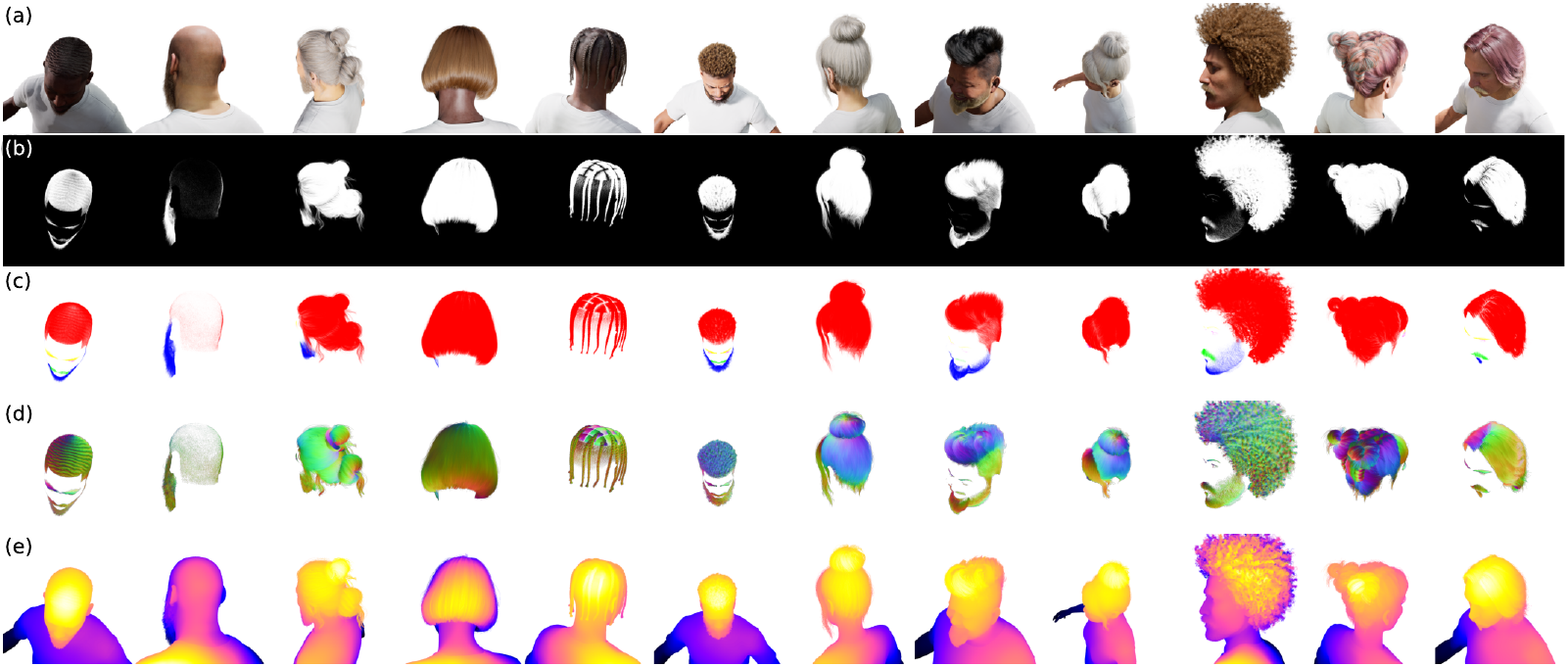}}
  \caption{\textbf{\datasetname Annotations}. We provide multiple labels for each image. (a) Rendered RGBA image of the MetaHuman. (b) Alpha matte for facial hair. (c) Soft segmentation labels for the 5 hair categories: red: scalp hair, yellow: eyebrows, pink: eyelashes, green: mustache, and blue: beard. (d) 3D hair growth direction. The axes are encoded as red: towards the screen, green: to the right, and blue: up. (e) Depth map of the entire body.}
  \label{fig:dataset_labels}
\end{figure*}

\section{Synthetic Hair Matting Dataset (\datasetname)}
\label{sec:dataset}
High-quality and realistic head models are essential for generating a detailed hair matting dataset in 4K image resolution.
We utilize the MetaHuman Creator 1.1.0~\cite{MetaHumanCreator} in combination with the Unreal Engine 5.6~\cite{UnrealEngine} to create high-quality and efficient renders of human faces.
The groom assets of MetaHuman models simulate individual hair strands, which results in highly detailed and realistic hair.
The final \datasetname dataset includes 43 MetaHuman models, each with a unique combination of groom assets, including 38 hairstyles, 18 eyebrows, 6 eyelashes, 16 mustaches, and 16 types of beards.
To further maximize the diversity of hair, we randomly switch groom and skin materials, randomize the camera pose, randomize the lighting, and animate the facial expressions by fitting the facial animation rig to random images of the AffectNet \cite{AffectNet} dataset.
For each image, we generate four types of annotations in 4K image resolution.
\begin{itemize}
    \item \textbf{Hair matte}: Continuous alpha matte for separating facial hair from the background.
    \item \textbf{Hair category}: Per-pixel soft-segmentation to categorize hair into the 5 classes of scalp hair, eyebrows, eyelashes, mustache, and beard.
    \item \textbf{Hair growth direction}: Per-pixel 3D normalized vectors that indicate the hair strands growth direction in the camera's coordinate system.
    \item \textbf{Depth}: Full body absolute depth map with a unit scale of $1$ cm, consistent with Unreal Engine's coordinate system.
\end{itemize}
A visualization of these labels is shown in \cref{fig:dataset_labels}.
We split the \datasetname dataset into 14.5k training samples and 1.5k test samples, ensuring that each MetaHuman appears in only one of the sets.

\begin{figure*}
  \centering{\includegraphics[width=\textwidth]{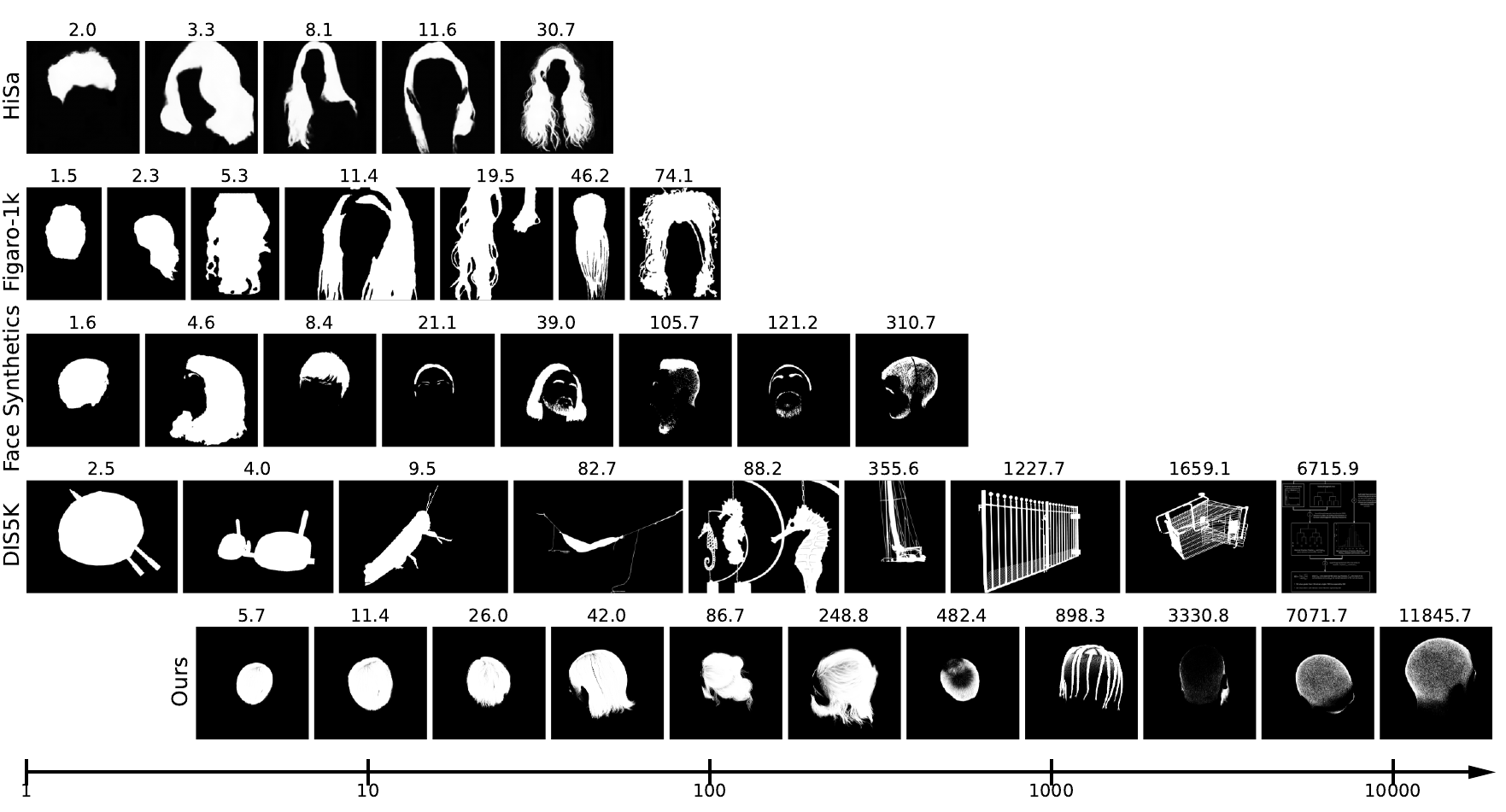}}
  \caption{\textbf{Annotations sorted by SIPQ}. We show a few randomly selected labels of hair segmentation and matting datasets ordered by their estimated SIPQ. Labels with low values on the left side consist of simple shapes, while annotations with high SIPQ on the right side show a high density of fine details. We also show examples of DIS5K~\cite{DIS5k}, as it has highly detailed and fine annotations.}
  \label{fig:sipq_examples}
\end{figure*}

\begin{figure*}
  \centering{\includegraphics[width=\textwidth]{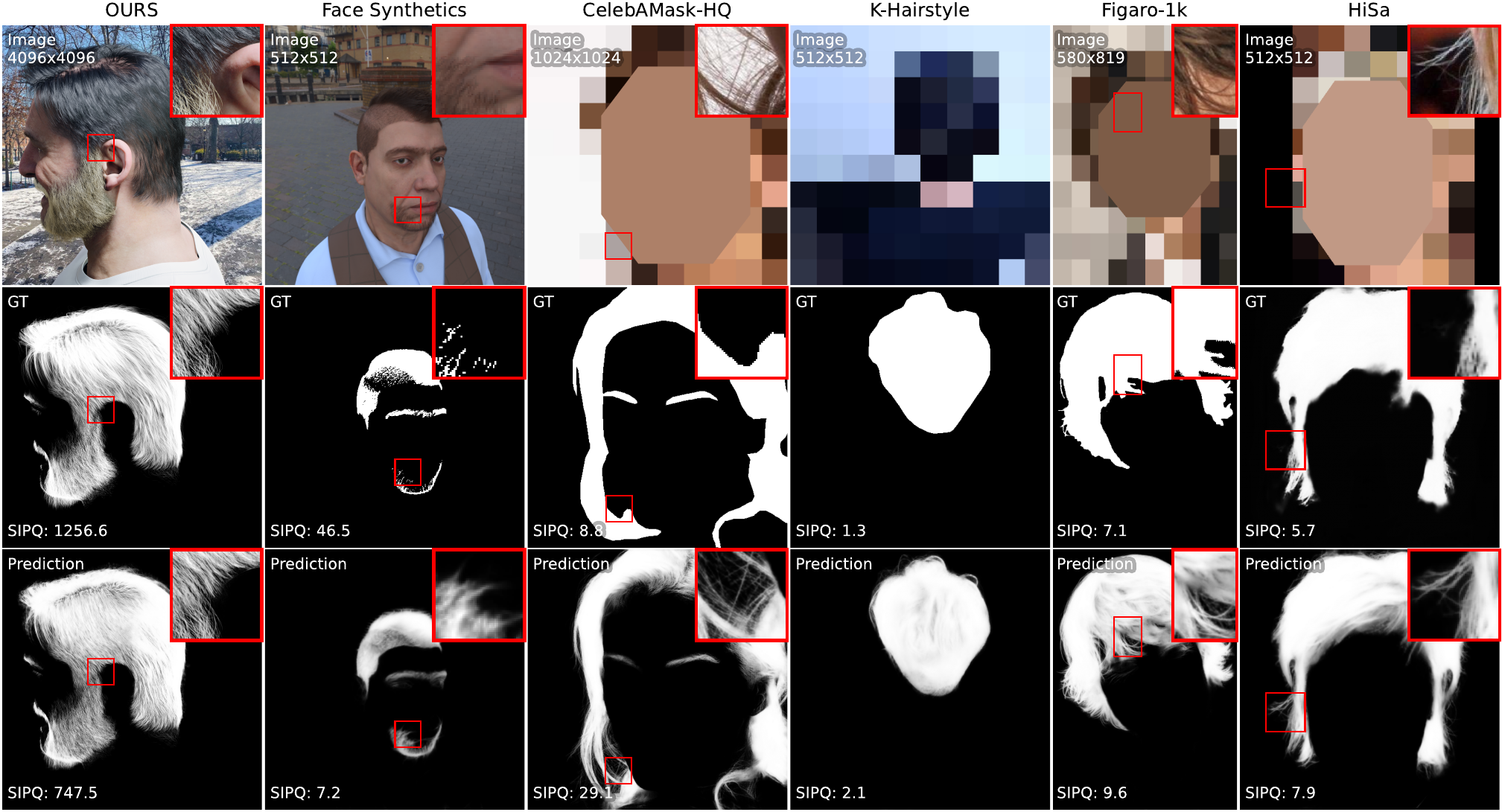}}
  \caption{\textbf{Qualitative comparison of dataset annotations}. We compare the quality of our hair matting labels with existing datasets. The last row shows predictions from our matting model trained on the \datasetname dataset.}
  \label{fig:datasets_matting_annotation_comparison}
\end{figure*}

\section{Measuring Label Complexity with SIPQ}
\label{sec:sipq_derivation}
In order to compare the annotation complexity of existing datasets with our own \datasetname labels, we require a metric that generalizes to both binary segmentation masks, as well as alpha mattes.
Prior work in dichotomous image segmentation \cite{DIS5k} utilizes the isoperimetric inequality quotient (IPQ) for measuring the object complexity:
\begin{align}
IPQ = \frac{L^2}{4 \pi A}
\label{eqn:ipq}
\end{align}
where $L$ denotes the perimeter length of the segmented object and $A$ its area.
However, this formulation is limited to binary segmentation masks, as it relies on explicitly estimating object contours.
We therefore formulate the Soft Isoperimetric Quotient (SIPQ) as:
\begin{align}
    SIPQ(I) = \frac{\left(\sum_{x, y} || \nabla I(x, y)||\right)^2}{4 \pi \sum_{x, y} ||I(x, y)||}
\end{align}
Here, $I$ is the soft segmentation map, $|| \nabla I(x, y)||$ is the gradient magnitude, and $\sum_{x, y} ||I(x, y)||$ describes the soft area.
This formulation is equal to the loss introduced in \cite{Liu2020ShapeawareMF}, but our purpose is not to utilize this as a compactness constraint.
Instead, we adapt it as a generalized measure of label complexity.
For a given binary segmentation mask $S$ with background pixels $0$ and foreground pixels $1$, the $SIPQ(S)$ approximates $IPQ(S)$ under the assumption that
\begin{align}
    L(S) \approx \sum_{x, y} || \nabla S(x, y)||
\end{align}
Intuitively, the SIPQ measures the amount of fine details proportional to the object's area.
We show visual examples ordered by their SIPQ in \cref{fig:sipq_examples} and an empirical comparison of IPQ and SIPQ in \cref{fig:siqp_vs_ipq} in the appendix.
We obtain the following properties of the proposed SIPQ measure:
\begin{itemize}
    \item \textbf{Uniform images}: Given an image $I$ with uniform values, we obtain $SIPQ(I) = 0$, which correctly reflects the absence of structural complexity. This is different from the original formulation of IPQ, which has a minimum value of $1$ for a circle.
    We choose to assign $SIPQ(I) = 0$ for the edge case of $I$ having only $0$ values, as this results in a division by zero.
    \item \textbf{Geometric invariance}: SIPQ is invariant to reflection, translation, and intensity scale. In its continuous formulation, it is also invariant under rotation and isotropic resizing.
\end{itemize}

\begin{table*}[h!]
    \caption{\textbf{Quantitative dataset comparison}. Comparison of our \datasetname dataset with other existing datasets for hair segmentation and hair matting. We additionally provide statistics of highly detailed dichotomous image segmentation (DIS) datasets as well as for the closely related task of portrait matting. For the K-Hairstyle~\cite{KHairstyle} dataset, we compute the estimates on a subset of $\sim$50k samples of the validation set. Synthetic datasets are marked with a $\star$.}
    \label{tab:quantiative_dataset_comparison}
    \centering
    \small
    \begin{tabular}{@{}l@{\hskip 6pt}llc@{\hskip 6pt}cr@{}}
        \toprule
        \multirow{2}{*}{Dataset} & \multirow{2}{*}{Label Type} & \multirow{2}{*}{Images}
        & \multicolumn{2}{c}{Annotation Dimension} 
        & \multirow{2}{*}{$SIPQ\uparrow$} \\
        & & 
        & $W \pm \sigma_W$ & $H \pm \sigma_H$\\
        \midrule
        ThinObject-5K~\cite{ThinObject5k}$^\star$ & DIS & $5748$ & $1325 \pm 959$ & $1186 \pm 910$ & $32.77$\\
        DIS5K~\cite{DIS5k} & DIS & $5470$& $\underline{3111} \pm 1360$ & $2513 \pm 1053$ & $130.68$\\
        P3M-10k~\cite{P3M10K} & Portrait Matting & $9921$ & $1348 \pm 290$ & $1321 \pm 264$ & $2.38$\\
        HHM2K~\cite{SparseMat} & Portrait Matting & $2000$ & $3570 \pm 1311$ & $4041 \pm 1526$ & $7.18$\\
        PPM-100~\cite{MODNet} & Portrait Matting & $100$ & $2997 \pm 1229$ & $\underline{2875} \pm 1209$ & $9.73$\\
        \arrayrulecolor{gray}\midrule\arrayrulecolor{black}
        K-Hairstyle~\cite{KHairstyle} & Hair Seg. & $500000$ & $512 \pm 0$ & $512 \pm 0$ & $3.26$ \\
        Figaro-1k~\cite{Figaro1k} & Hair Seg. & $1050$ & $635 \pm 214$ & $718 \pm 191$ & $5.33$\\
        CelebAMask-HQ~\cite{CelebAMask-HQ} & Hair Seg. & $30000$ & $512 \pm 0$ & $512 \pm 0$ & $5.79$ \\
        Face~Synthetics~\cite{FaceSynthetics}$^\star$ & Hair Seg. & $100000$ & $512 \pm 0$ & $512 \pm 0$ & $20.75$\\
        HiSa~\cite{HairStep} & Hair Matting & 1250 & $512 \pm 0$ & $512 \pm 0$ & $7.08$\\
        \datasetname 512x512 (Ours)$^\star$ & Hair Matting & 16000 & $512 \pm 0$ & $512 \pm 0$ & $22.47$\\
        \datasetname 1024x1024 (Ours)$^\star$ & Hair Matting & 16000 & $1024 \pm 0$ & $1024 \pm 0$ & $88.48$\\
        \datasetname 2048x2048 (Ours)$^\star$ & Hair Matting & 16000 & $2048 \pm 0$ & $2048 \pm 0$ & $\underline{325.41}$\\
        \textbf{\datasetname (Ours)}$^\star$ & Hair Matting & 16000 & $\mathbf{4096} \pm 0$ & $\mathbf{4096} \pm 0$ & $\mathbf{734.75}$\\
        \bottomrule
    \end{tabular}
\end{table*}
% Dichotomous Image Seg

\begin{table*}[h!]
    \caption{\textbf{Supported hair categories across datasets}. We compare the available hair annotations by category. Our dataset is the only one which provides labels for all five types of facial hair.}
    \label{tab:hair_categories}
    \centering
    \small
    \begin{tabular}{@{}llc@{\hskip 4pt}c@{\hskip 4pt}c@{\hskip 4pt}c@{\hskip 4pt}c@{}}
        \toprule
        \multirow{2}{*}{Dataset} & \multirow{2}{*}{Images} & \multicolumn{5}{c}{Hair Categories}
        \\
        & & Scalp & Brows & Lashes & Stache & Beard\\
        \midrule
        CelebAMask-HQ~\cite{CelebAMask-HQ} & $30000$ & {\color{ForestGreen}\cmark} & {\color{ForestGreen}\cmark} & {\color{BrickRed}\xmark} & {\color{BrickRed}\xmark} & {\color{BrickRed}\xmark} \\
        Face~Synthetics~\cite{FaceSynthetics}$^\star$ & $100000$ & {\color{ForestGreen}\cmark} & {\color{ForestGreen}\cmark} & {\color{BrickRed}\xmark} & {\color{ForestGreen}\cmark} & {\color{ForestGreen}\cmark} \\
        K-Hairstyle~\cite{KHairstyle} & $500000$ & {\color{ForestGreen}\cmark} & {\color{BrickRed}\xmark} & {\color{BrickRed}\xmark} & {\color{BrickRed}\xmark} & {\color{BrickRed}\xmark} \\
        Figaro-1k~\cite{Figaro1k} & $1050$ & {\color{ForestGreen}\cmark} & {\color{BrickRed}\xmark} & {\color{BrickRed}\xmark} & {\color{BrickRed}\xmark} & {\color{BrickRed}\xmark} \\
        HiSa~\cite{HairStep} & 1250 & {\color{ForestGreen}\cmark} & {\color{BrickRed}\xmark} & {\color{BrickRed}\xmark} & {\color{BrickRed}\xmark} & {\color{BrickRed}\xmark} \\
        \textbf{\datasetname (Ours)}$^\star$ & 16000 & {\color{ForestGreen}\cmark} & {\color{ForestGreen}\cmark} & {\color{ForestGreen}\cmark} & {\color{ForestGreen}\cmark} & {\color{ForestGreen}\cmark} \\
        \bottomrule
    \end{tabular}
\end{table*}

% --------------------------------------
% Data Analysis
% --------------------------------------
\section{Data Analysis} \label{sec:data_analysis}
\paragraph{Qualitative comparison}
We compare our \datasetname dataset with previous work and show that the provided level of detail of existing datasets is not sufficient to identify individual hair strands accurately.
\cref{fig:datasets_matting_annotation_comparison} shows images of various datasets and their ground truth annotations.
As can be seen in the left-most column, our dataset provides a significantly more detailed annotation with individual hairs being clearly distinguishable.
In the second column, we can observe one major issue of providing only binary segmentation labels.
The beard's hair strands are too small to be accurately represented by the binary segmentation mask, and therefore, we obtain a very scattered and noisy ground truth.
In columns 3 to 6, we notice that the ground truth annotation is missing a lot of fine details, which are essential for training a precise hair matting model.
For comparison, the last row shows predictions from our \modelname model, trained on our \datasetname dataset, which exhibits a substantially higher level of detail than the provided ground-truth labels.

\paragraph{Empirical comparison}
We compare our dataset with previous work in \cref{tab:quantiative_dataset_comparison} and utilize the derived SIPQ to empirically compare our annotations with those provided by other datasets.
Our dataset provides the highest resolution annotations among all datasets and achieves a significantly higher average SIPQ of $734.75$, indicating the presence of fine details and structures in our alpha mattes.
Most importantly, a high image resolution does not necessarily imply a higher density of details.
In the task of portrait matting, all three datasets have a lower SIPQ than $10$.
For instance, PPM-100~\cite{MODNet} has an average SIPQ of only $9.73$, despite the high resolution of approximately $3{,}000 \times 3{,}000$ pixels.
This highlights a significant challenge of manually annotating hair mattes in high-resolution images.
For portrait images, increasing the image resolution mainly results in changes to the body's silhouette, while hair matting in higher resolutions exposes more details in regions with thin hair.
To further analyze this effect, we also list three downsampled versions of our \datasetname dataset.\footnote{Images were downsampled using area-based interpolation to prevent the introduction of aliasing artifacts and high-frequency noise.}
As expected, SIPQ decreases with lower resolution, reflecting the loss of high-frequency detail.
When compared with Face Synthetics~\cite{FaceSynthetics} at a matched resolution of $512 \times 512$, our dataset exhibits comparable SIPQ values, while the full 4K version has approximately 30x finer detail.
In \cref{tab:hair_categories}, we provide a comparison of different hair categories covered by each dataset. 
To the best of our knowledge, our dataset is the first to provide detailed eyelash annotations for full upper body portrait images, due to the fine structures of lashes that are typically only visible at high resolutions.

\begin{table*}[h!]
    \caption{\textbf{Impact of training resolution}. We show the impact of training \modelname on different image resolutions of our \datasetname dataset.}
    \label{tab:model_vs_dataset_resolutions}
    \centering
    \small
    \begin{tabular}{@{}lc@{\hskip 4pt}cc@{\hskip 4pt}cc@{\hskip 4pt}cc@{\hskip 4pt}c@{}}
        \toprule
        \modelname & \multicolumn{8}{c}{\datasetname} \\
        Training & \multicolumn{2}{c}{$512 \times 512$} & \multicolumn{2}{c}{$1024 \times 1024$} & \multicolumn{2}{c}{$2048 \times 2048$} & \multicolumn{2}{c}{$4096 \times 4096$}\\
        Resolution & $GRAD\downarrow$ & $MSE\downarrow$ & $GRAD\downarrow$ & $MSE\downarrow$ & $GRAD\downarrow$ & $MSE\downarrow$ & $GRAD\downarrow$ & $MSE\downarrow$\\
        \midrule
        $512 \times 512$ & $\mathbf{0.0260}$ & $\mathbf{0.0017}$ & $0.0248$ & $0.0019$ & $0.0243$ & $0.0024$ & $0.0190$ & $0.0029$\\
		$1024 \times 1024$ & $\underline{0.0274}$ & $\underline{0.0018}$ & $\mathbf{0.0213}$ & $\mathbf{0.0015}$ & $\underline{0.0197}$ & $0.0017$ & $0.0157$ & $0.0021$\\
		$2048 \times 2048$ & $0.0331$ & $0.0022$ & $\underline{0.0223}$ & $\underline{0.0016}$ & $\mathbf{0.0175}$ & $\mathbf{0.0014}$ & $\underline{0.0130}$ & $\underline{0.0016}$\\
        $4096 \times 4096$ & $0.0398$ & $0.0028$ & $0.0274$ & $0.0020$ & $0.0199$ & $\underline{0.0015}$ & $\mathbf{0.0106}$ & $\mathbf{0.0013}$\\
        \bottomrule
    \end{tabular}
\end{table*}

\begin{table*}
    \caption{\textbf{Domain generalization}. We compare our predicted mattes with real-data annotations. For \modelname trained on matting labels, we set the foreground thresholds to \datasetname: $0.4$, and HiSa:$0.6$.}
    \label{tab:domain_generalization}
    \centering
    \small
    \begin{tabular}{@{}l@{\hskip 6pt}lc@{\hskip 3pt}c@{\hskip 3pt}cc@{\hskip 3pt}c@{\hskip 3pt}cc@{\hskip 3pt}c@{\hskip 3pt}c@{}}
        \toprule
        \multirow{2}{*}{Model} & \multirow{2}{*}{Training Dataset} & \multicolumn{3}{c}{Figaro-1k} & \multicolumn{3}{c}{K-Hairstyles} & \multicolumn{3}{c}{CelebAMask-HQ}\\
        & & ACC$\uparrow$ & mIoU$\uparrow$ & F1$\uparrow$ & ACC$\uparrow$ & mIoU$\uparrow$ & F1$\uparrow$ & ACC$\uparrow$ & mIoU$\uparrow$ & F1$\uparrow$\\
        \midrule
        \cite{TwoStageHumanHairSegmentation} & Figaro-1k & $97.2$ & $91.3$ & $95.2$  & - & - & - & - & - & - \\
        \cite{FineGrainedHumanHairSegmentation} & Figaro-1k & $\underline{98.0}$ & $\textbf{95.4}$ & $\textbf{98.3}$ & - & - & - & - & - & - \\
        PAN \cite{PAN} & K-Hairstyle & - & - & - & $\textbf{98.6}$ & $\textbf{91.9}$ & $\textbf{95.7}$ & - & - & - \\
        \cite{FineGrainedHumanHairSegmentation} & CelebAMask-HQ & - & - & - & - & - & - & $95.3$ & $\textbf{92.6}$ & $\textbf{97.8}$\\
        H2M2 & Figaro-1k & $\mathbf{98.2}$ & $\underline{94.9}$ & $\underline{97.3}$ & $98.3$ & $91.4$ & $95.0$ & $\underline{96.5}$ & $86.4$ & $91.3$\\
		H2M2 & HiSa & $95.7$ & $87.9$ & $92.4$ & $\underline{98.4}$ & $\textbf{91.9}$ & $\underline{95.3}$ & $\underline{96.5}$ & $86.3$ & $91.2$\\
        H2M2 & CelebAMask-HQ & $95.8$ & $88.2$ & $92.7$ & $98.3$ & $91.6$ & $95.1$ & $\textbf{97.6}$ & $\underline{90.7}$ & $\underline{93.2}$\\
		H2M2 & \datasetname (Ours) & $96.8$ & $90.4$ & $94.5$ & $98.2$ & $91.1$ & $94.8$ & $95.8$ & $83.3$ & $89.4$\\
        \bottomrule
    \end{tabular}
\end{table*}

% --------------------------------------
% Experiments
% --------------------------------------
\section{Experiments}
\label{sec:experiments}
In order to analyze the domain gap and gain of our precise 4K matte annotations, we train the same baseline human hair matting model (\modelname) inspired by the recent architecture DAViD~\cite{DAViD} on \datasetname, CelebAMask-HQ, Figaro-1k, and HiSa.
DAViD is a dense prediction transformer (DPT)~\cite{DPT} that resizes input images to a fixed resolution for the transformer backbone while maintaining the original resolution for the dense prediction head.
This design allows models trained on lower-resolution data to scale more effectively to high-resolution images, enabling a fairer comparison of models trained on low-resolution real-world datasets vs. our high-resolution synthetic data.
For our \datasetname dataset, we train two models, \modelname-Matte predicts the alpha matte of all facial hair, while \modelname-Attributes estimates hair category, orientation, and depth.
Whenever we require predictions of only a specific hair category, we use the \modelname-Attributes' per-pixel category prediction to mask undesired regions from the \modelname-Matte prediction.
A more detailed explanation of the models' architectures and training schedules is provided in the appendix.

\paragraph{Impact of training resolution}
In \cref{tab:model_vs_dataset_resolutions}, we analyze the impact of the training resolution on inference at higher hair matting resolutions.
While our model, as per design, exhibits strong generalization across different scales, we observe a clear improvement when training at higher resolutions.
These results indicate that high-quality hair matting at 4K image resolution requires a high-resolution corresponding training set.

\paragraph{Domain generalization}
Quantitatively evaluating our 4K hair-matting model on real data remains challenging, as no annotated 4K dataset currently exists, and manually labeling such high-resolution images is practically infeasible. 
As a result, we rely on lower-resolution datasets for our evaluation in \cref{tab:domain_generalization}.
While these datasets do not fully reflect the performance of our 4K model, they provide a rough estimate of the domain gap.
We additionally list results of related work for reference.
On Figaro-1k and K-Hairstyle, we achieve accuracies above 95\% and mIoU scores above 90\%, indicating good generalization from synthetic to real data.
Importantly, when comparing cross-dataset performances, we note that training on \datasetname achieves higher mIoU on Figaro-1k than training on the HiSa or CelebAMask-HQ dataset.
On CelebAMask-HQ, we observe a larger domain gap with a mIoU of around 83\%.

\begin{figure*}
  \centering{\includegraphics[width=\textwidth]{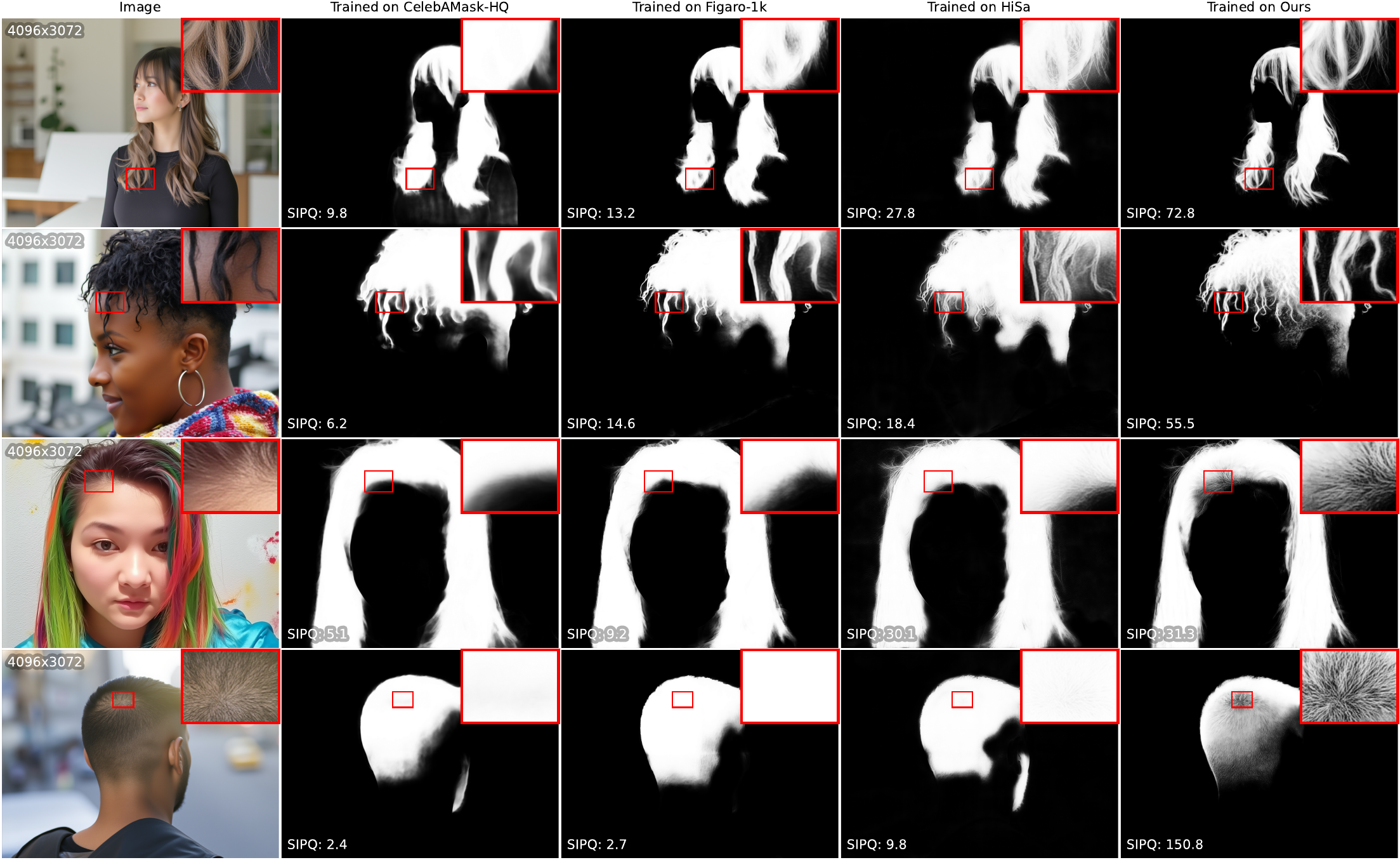}}
  \caption{\textbf{High-resolution results}. We compare scalp hair predictions of \modelname when trained on different datasets. The model trained on our dataset separates fine hair details from the background significantly better. Images are synthetically generated with Diffusion-4K~\cite{Diffusion4K}.}
  \label{fig:train_on_real_highres}
\end{figure*}

\begin{figure}[!ht]
  \centering
  \includegraphics[width=\columnwidth]{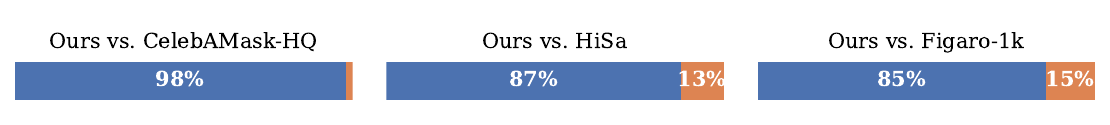}
  \caption{\textbf{A/B Test Results}. Comparing \modelname-Matte trained on our \datasetname against the model trained on other datasets. All models only predict the scalp hair matte to enable a fair comparison.}
  \label{fig:ab_test}
\end{figure}

\paragraph{Results on high resolution images}  
\cref{fig:train_on_real_highres} compares our \modelname model trained on the 4K-resolution \datasetname dataset with models trained on low-resolution real datasets.
To avoid exposing personal data, we utilize synthetically generated images with Diffusion-4K~\cite{Diffusion4K}.
We only consider matting the scalp hair, as these labels are provided across all four datasets.
While all models produce similar coarse predictions, we observe a clear loss of fine details in models trained on low-resolution real data.
The model trained on \datasetname cleanly separates individual strands, whereas models trained on the other datasets fail to do so.
This observation is further reflected by a substantial increase in SIPQ values.
In the last row, models trained on CelebAMask-HQ~\cite{CelebAMask-HQ} and HiSa~\cite{HairStep} also incorrectly classify the beard as part of the scalp hair matte.
This is likely because these datasets primarily consist of frontal facial views.
As we do not have ground-truth 4K hair matting labels for real data, we perform an A/B test on 30 randomly selected images of the HHM2K~\cite{SparseMat} dataset with 15 users.
The test results are shown in \cref{fig:ab_test}.
Users favored alpha mattes generated by models trained on \datasetname in over $80\%$ of cases compared to those trained on existing real-world datasets.
This preference was most pronounced against CelebAMask-HQ ($98\%$), which is likely because of its coarse binary segmentation masks.

\begin{figure}[!ht]
  \centering
  \includegraphics[width=\columnwidth]{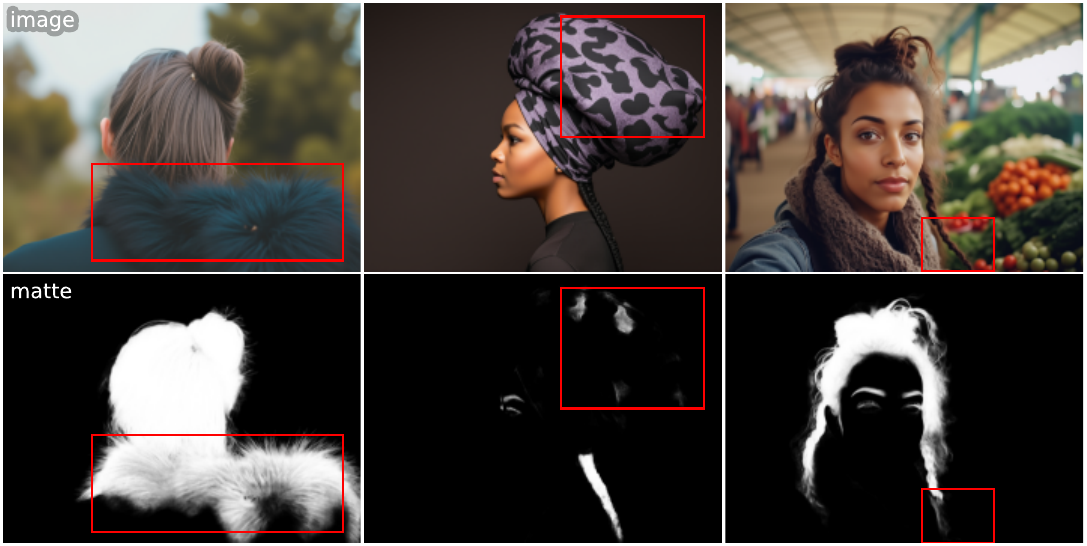}
  \caption{\textbf{Failure Cases}. Examples of failure cases of training \modelname on \datasetname, tested on synthetic images generated by Diffusion-4K~\cite{Diffusion4K}.}
  \label{fig:failure_cases}
\end{figure}

\section{Discussion and Limitations}
\paragraph{SIPQ} While the SIPQ is a useful tool for measuring the complexity of ground truth annotations, it does not distinguish between different types of structures.
Due to the use of image gradients, the SIPQ is sensitive to high-frequency noise.
This can be partly observed in \cref{fig:datasets_matting_annotation_comparison}, where the noisy regions of the Face Synthetics' ground truth result in a high SIPQ.
The issue is mainly present in synthetic segmentation datasets in areas that are close to the binary segmentation threshold.
In our \datasetname dataset, we mitigate this issue by providing alpha mattes and utilizing multi-sampled anti-aliasing during label generation.
\cref{tab:quantiative_dataset_comparison} we also observe that area-based downsampling of \datasetname from 4K to 2K image resolution results in a SIPQ of $325.41$, further demonstrating that the large SIPQ does not originate from high frequency label noise.

\paragraph{\datasetname} The dataset does not incorporate randomized clothing or accessories, which limits the diversity of non-hair appearances during training.
We partially address this by compositing background images from the clothing and accessories dataset~\cite{ClothingDataset}. Despite our efforts, as shown in \cref{fig:failure_cases}, we still observe that this limitation occasionally leads to accessories or fur being wrongly predicted as hair.
We also noticed that hairstyles like braids or plaits might also lead to prediction errors, as illustrated in the rightmost example, as they might not be as well-represented in the dataset.
\section{Conclusion}
\label{sec:conclusion}
In this work, we introduced \datasetname, a novel synthetic hair matting dataset consisting of 16,000 RGBA images at an ultra-high 4K resolution, created using the Unreal Engine and MetaHuman Creator.
By adapting the soft isoperimetric inequality quotient (SIPQ) as a metric for label complexity, we demonstrated the significant increase in structural detail present at higher resolutions of the hair matting task.
Finally, our experiments show that models trained on our new synthetic dataset generalize well to real-world images, often exceeding the quality of models trained on existing datasets and enabling highly detailed 4K hair matting.
\clearpage
\begin{ack}
\textbf{Funding Transparency Statement} \\
Markus Karmann is supported by a PhD sponsorship from vivo.

\textbf{Acknowledgements} \\
Applied MetaHuman powered synthetic data from Epic Games.
Contact Epic Games for licensing options. 
\end{ack}

\bibliographystyle{unsrtnat}
\bibliography{main}

%%%%%%%%%%%%%%%%%%%%%%%%%%%%%%%%%%%%%%%%%%%%%%%%%%%%%%%%%%%%

\clearpage
\appendix

\section{Appendix / Supplementary Material}

% -----------------------------------------
% Implementation Details of Data Generation
% -----------------------------------------
\subsection{Implementation Details of the Data Generation}
\label{sec:dataset_generation_implementation_details}

\begin{figure}
  \centering{\includegraphics[width=0.9\textwidth]{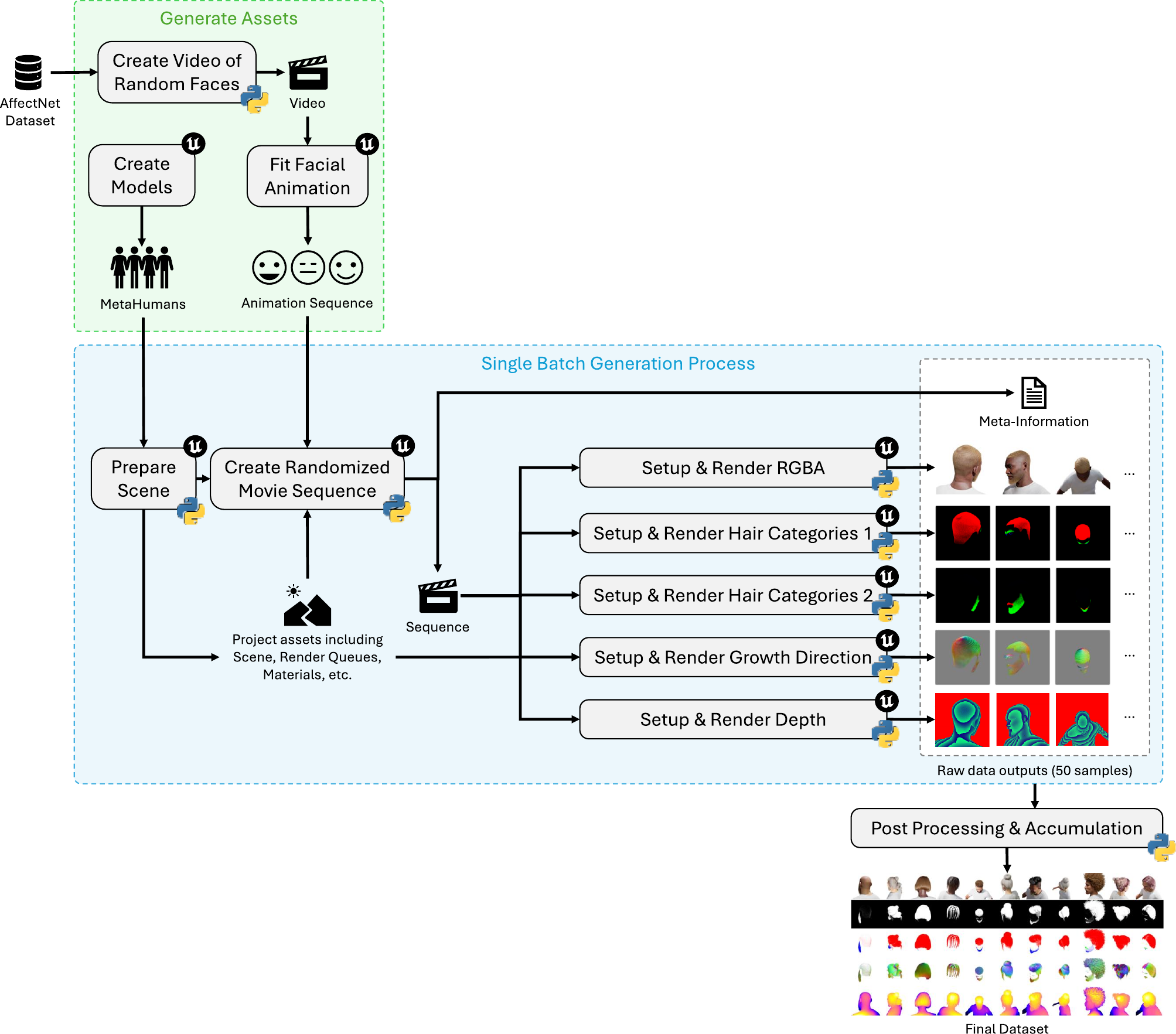}}
  \caption{\textbf{Overview of data generation pipeline}. This figure shows an overview of our data generation pipeline. Automated steps are marked with a Python icon. Steps involving the Unreal Engine~\cite{UnrealEngine} and MetaHuman Creator~\cite{MetaHumanCreator} are marked with an Unreal Engine icon. The MetaHuman models and facial animation sequence are only created initially. The \textit{Single Batch Generation Process} is repeated until we obtain the desired number of samples for our dataset.}
  \label{fig:pipeline}
\end{figure}

\begin{figure}
  \centering{\includegraphics[width=\columnwidth]{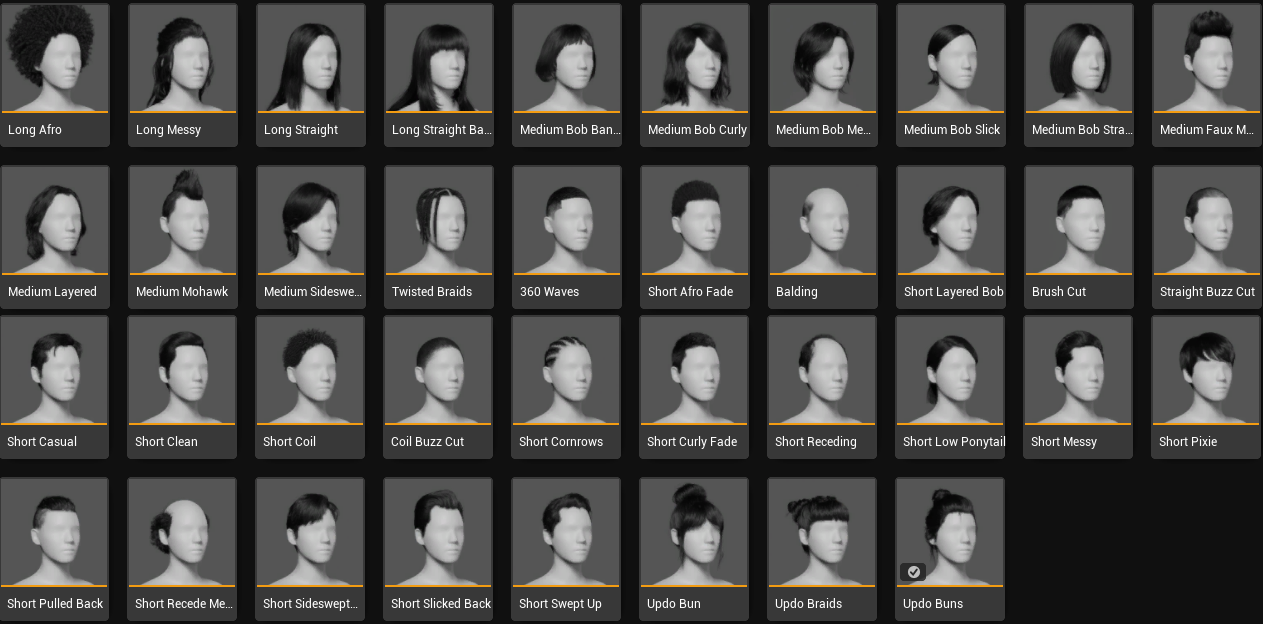}}
  \centering{\includegraphics[width=\columnwidth]{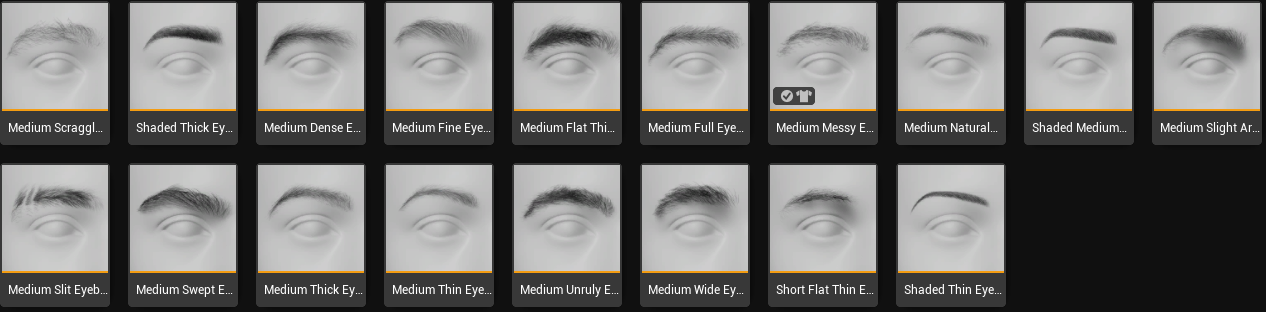}}
  \centering{\includegraphics[width=\columnwidth]{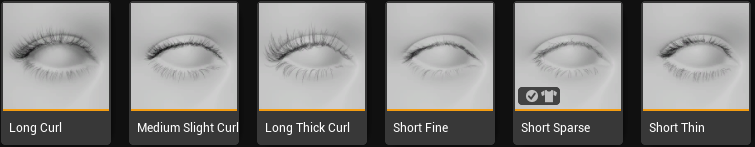}}
  \centering{\includegraphics[width=\columnwidth]{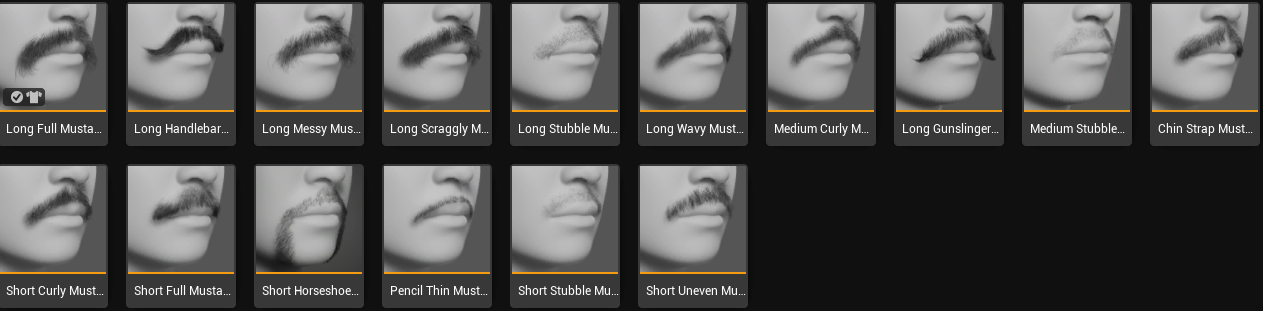}}
  \centering{\includegraphics[width=\columnwidth]{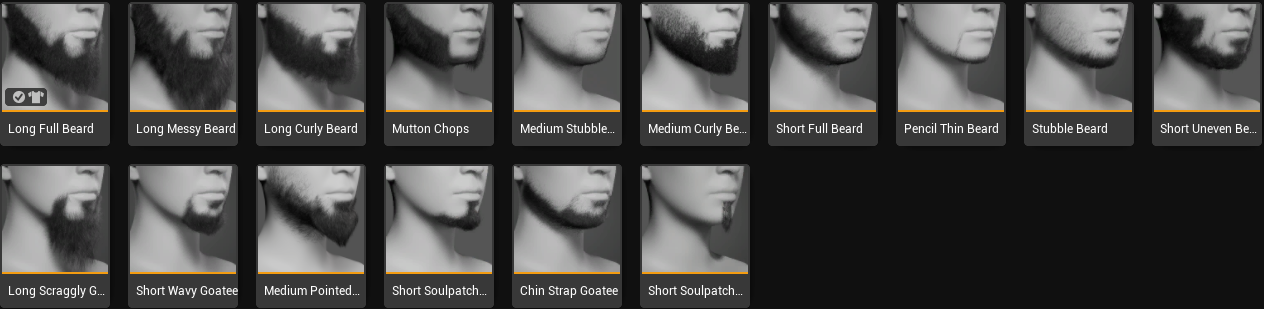}}
  \caption{\textbf{Groom assets}. Visualization of all groom assets grouped by their category.}
  \label{fig:groom_asstes}
\end{figure}

\paragraph{Creating assets}
\cref{fig:pipeline} shows an overview of our proposed data generation pipeline.
We use the MetaHuman Creator~\cite{MetaHumanCreator} and Unreal Engine 5.6~\cite{UnrealEngine} to create our synthetic dataset.
The MetaHuman Creator is fully integrated in the Unreal Engine since version 5.6, but can not yet be accessed by the Unreal Engine Python API.
We therefore create 43 MetaHuman models manually, which takes about 20 minutes per model due to the rigging and texturing process.
The resulting assets consist of the 29 presets provided by the MetaHuman Creator and 14 additional models to cover all available 38 hairstyles, 18 eyebrows, 6 eyelashes, 16 mustaches, and 16 beard groom assets.
We visualize a collection of all groom assets in \cref{fig:groom_asstes}, especially the scalp hair, which provides a large variety of hairstyles.

\begin{figure}
  \centering{\includegraphics[width=\textwidth]{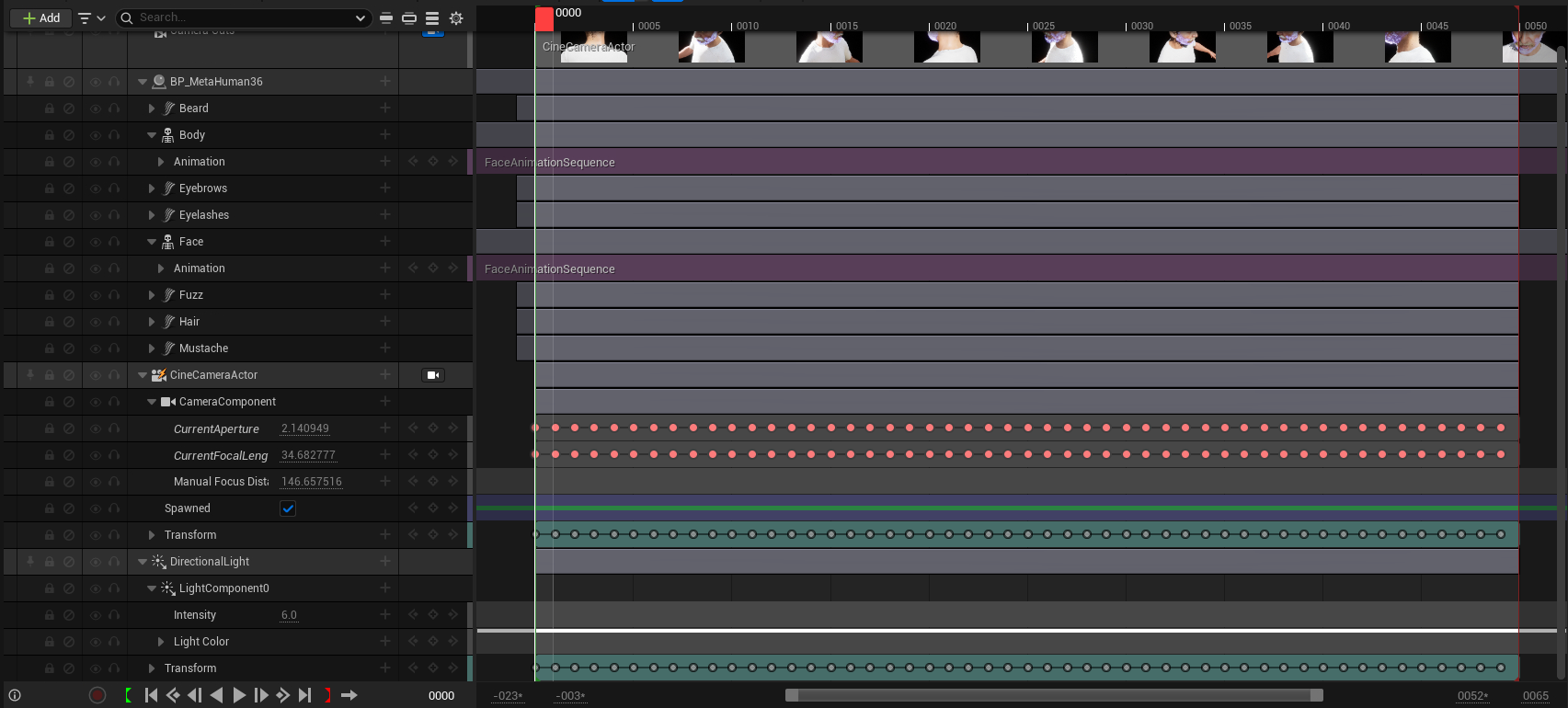}}
  \caption{\textbf{Sequencer Tracks}. A screenshot of the Unreal Engine Sequencer showing all animation tracks that are used for generating the dataset. All tracks are automatically generated via the Unreal Engine Python API.}
  \label{fig:sequencer_tracks}
\end{figure}

\paragraph{General Setup}
We use the Unreal Engine Sequencer and Render Queue to create images and labels.
This process is entirely automated with the Unreal Engine Python API and runs on a single system with a NVIDIA GeForce RTX 4090.
Generating the 16k samples dataset on this workstation takes about 10 days with ~50 seconds per sample.
We use about $10$\% of the 16k rendered images for our test set, resulting in 14.5k training samples and 1.5k test samples.
The test set has 4 unique MetaHuman models with scalp hair assets that are not included in the training set.
Images and labels are rendered as short sequences in batches of 50 frames.
We first run a Python script to randomize the MetaHuman, scene, and camera parameters and save these changes in the level and sequence assets.
A screenshot of a generated 50-frame movie sequence is shown in \cref{fig:sequencer_tracks}.

\paragraph{Obtaining Labels}
Each type of label requires a combination of switching materials in the scene and using post-processing materials.
In order to prevent potential changes from carrying over from one label generation process to another, we close and restart the Unreal Engine after each label type is finished without saving the changes.
\begin{itemize}
    \item \textbf{RGBA}: Uses normal Deferred Renderer with transparent scene background.
    \item \textbf{Depth}: Uses Unlit Deferred Renderer with a custom post-processing material to clip large depth values. In early tests, we noticed that running Unreal Engine on Linux applies post processing floating point ouputs, and we therefore are not able to obtain raw floating-point depth values straight from the render output. Fortunately, this effect does not appear in uint8 RGB outputs, and we therefore use the custom postprocessing material to first encode the depth into 24-bit RGB (see raw depth output in \cref{fig:pipeline}).
    \item \textbf{Hair Matte}: Uses Unlit Deferred Renderer and exchanges every material and background to white except the hair, which is rendered in black. MetaHumans also simulate very tiny hair on skin, called fuzz, which we do not include in our labels, as they are mostly indistinguishable from human skin.
    \item \textbf{Hair Growth Direction}: Uses a custom hair material that maps the hair attribute tangent vector to the base color buffer. We then use the unlit deferred renderer and use an additional post-process material to write the BaseColor buffer to images.
    \item \textbf{Hair Categories}: Setting every material and background black and assigning Scalp Hair, Eyelashes, and Eyebrows each to a RGB channel for the base color buffer. This is done in 2 passes. The second renders the mustache and beard hair.
\end{itemize}
In total, we run the Unreal Engine 7 consecutive times for rendering a single batch of 50 images.
We show the raw output of these renders in \cref{fig:pipeline}.
Some of the outputs require additional post-processing due to gamma correction, which might result in small errors of about $2\%$ for smaller alpha values.

\paragraph{Hair Physics}
Unreal Engine provides hair physics simulations, but unfortunately, we are not able to achieve consistent results across multiple render passes.
The simulation is not deterministic and slightly varies after a few steps, which leads to misaligned alpha mattes and RGBA renders.
Caching the Niagara hair simulation system is unfortunately not possible, as reloading the project creates new instances of Niagara, which are disconnected from the existing cache in the sequencer.
We therefore deactivate the hair simulation and reset the hair strands to their initial state at each frame.

\paragraph{Scene lighting}
Other works \cite{FaceSynthetics} use skyboxes or similar environment maps to achieve a realistic and diverse lighting setup.
As we are using the deferred renderer, most environment maps result in undesired artifacts.
These artifacts are especially present in the hair.
We therefore use a single directional light as our main light source.
We randomize the direction of the light for each frame.

\paragraph{Facial animation}
In order to have a large variety of facial expressions, we select 1000 random images of the AffectNet~\cite{AffectNet} database and fit the MetaHuman facial animation control rig to these images.
We interpolate between different expressions to obtain a facial animation of 10000 frames in total.
A random section of this \textit{Animation Sequence} is used for each batch to keep facial expressions diverse and consistent across multiple renders.

\paragraph{Camera setup}
Compared to existing hair segmentation datasets, we provide a wide range of camera angles.
The camera's position is sampled randomly around the MetaHuman's head while ensuring that the head is always in the center of the frame.
We also vary the camera distance, aperture, and focal length.

\paragraph{Hair material}
We randomize the hair material before rendering each image batch. 
This is done by either selecting an existing material from other MetaHuman hair assets or setting the material properties of color and roughness to random values.

\paragraph{Maximizing image quality}
We use Unreal Engine's deferred renderer, as this method is a lot faster and requires less memory then ray tracing while maintaining very good visual quality.
In order to render in 4K image resolution, we split the images into $4 \times 4$ patches of $1024 \times 1024$.
We deactivated effects such as motion blur, depth of field, and temporal anti-aliasing to prevent image artifacts and improve picture clarity.
Not relying on temporal anti-aliasing is necessary, as randomly setting the camera's position and orientation, as well as the direction of the light source, causes significant changes between each frame.
We therefore set the spatial anti-aliasing to 64 samples for RGBA and alpha rendering to improve the quality of edges and thin hair strands.
For rendering depth and hair growth direction, the spatial anti-aliasing is set to 1 sample per pixel.
This ensures that we do not obtain artifacts of hair attributes mixing with the background.
For example, thin hair strands in front of the transparent background would result in mixed, very distant depth values.

\begin{figure*}
  \centering{\includegraphics[width=\textwidth]{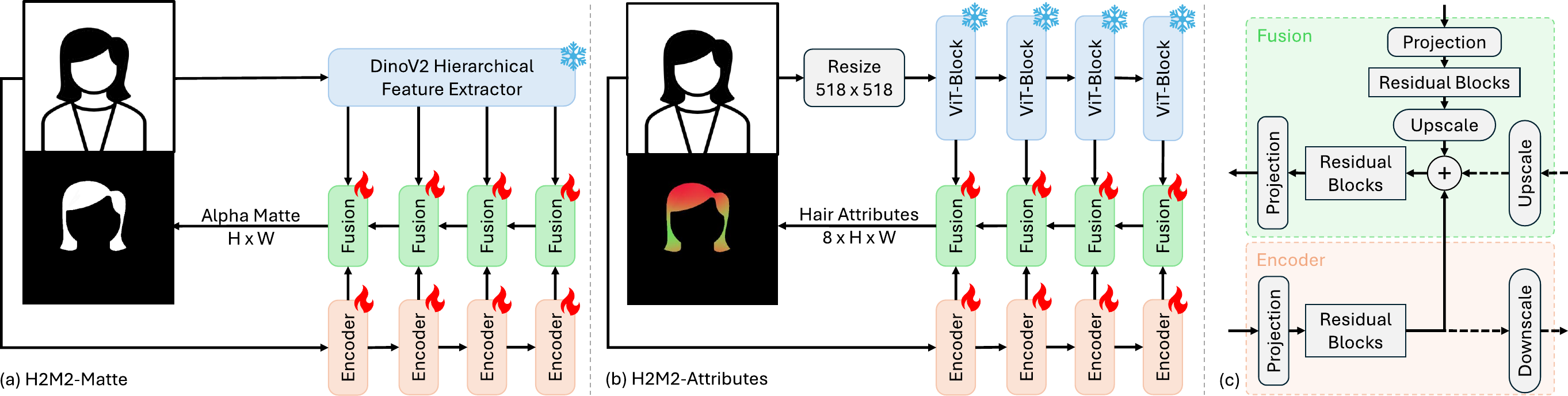}}
  \caption{\textbf{Overview of model architectures}. (a): \modelname-Matte architecture for alpha matte prediction. (b): \modelname-Attributes used to predict hair category, orientation, and depth. (c): The structure of one Encoder and one Fusion block. We mark model parts that are frozen in the training stage with a snowflake and parts that are updated with a flame.}
  \label{fig:model_architecture}
\end{figure*}

%\begin{figure}[!ht]
%  \centering
%  \includegraphics[width=0.5\columnwidth]{figures/hiararchical_feature_encoder.pdf}
%  \caption{\textbf{DinoV2 Hierarchical Feature Encoder}. Increasing spatial density of DinoV2 features by utilizing 3 hierarchies of patches.}
%  \label{fig:hierarchical_feature_encoder}
%\end{figure}

% -------------------------------
% Implementation Details of Model
% -------------------------------
\subsection{Implementation Details of \modelname}
\label{sec:model_and_training_implementation_details}
We implement all models in Python using PyTorch for GPU-accelerated training and image processing. None of the trainable components use layer normalization or batch normalization.
Both \modelname-Matte and \modelname-Attributes share closely related architectures, combining features extracted from the DINOv2~\cite{DINOv2} transformer, with a high-resolution ResNet inspired by \cite{SparseMat}.

\paragraph{\modelname-Matte} 
\modelname-Matte uses a hierarchical feature extractor and produces a single output channel. 
The ResNet configuration is configured to a depth of 8 and a width of 8.

\paragraph{\modelname-Attributes} 
\modelname-Attributes uses intermediate feature representations of DinoV2 and outputs eight channels in total.
The ResNet width of 12 and a shallower depth of 4 are chosen to accommodate the increased number of output channels.
For \modelname-Attributes, all non-hair pixels are labeled as undefined during training and excluded from the loss computation in order to simplify the learning objective.
At inference time, the predicted matte from \modelname-Matte enables us to mask out these undefined regions.

\paragraph{High-resolution ResNet}
The high-resolution \textit{Encoder} and \textit{Fusion} blocks are constructed using convolutions exclusively, enabling us to predict directly at the native image resolution without any resizing.\footnote{The only requirement is that the image dimensions are divisible by 8, which we can ensure by applying padding to the input.}
This adaptive design is crucial, as resizing the input and output of our model results in inaccuracies for small pixel-sized hair strands.
Compared to DPT~\cite{DPT}, both \textit{Encoder} and \textit{Fusion} blocks use multiple residual blocks instead of a single residual unit, such that the overall structure resembles a ResNet~\cite{ResNet}.
This is similar to the high-resolution module of previous 4K portrait matting works, such as~\cite{SparseMat}.
We set the final output channel dimension to 1 for \modelname-Matte and 8 for \modelname-Attribute.

\paragraph{Low-resolution Transformer}
For the low-resolution backbone, we use the pre-trained Vision Transformer ViT-S from DINOv2~\cite{DINOv2}.
Following \cite{DAViD}, input images are first resized to a fixed resolution, which improves generalization across varying image sizes.
For \modelname-Attributes, we extract intermediate features following \cite{DepthAnythingV2}, as these provide richer spatial context beneficial for attribute prediction.
In contrast, \modelname-Matte uses the output “all-purpose” features of DinoV2~\cite{DINOv2}, which are already sufficiently informative and expected to generalize well from synthetic to real data.
Inspired by \cite{MVANet}, the input image is partitioned into multiple hierarchical patch levels to extract multi-scale features, which are then reassembled into a dense feature map, preserving fine spatial details.
Additional information regarding the model architecture and implementation is provided in the supplementary material.

\paragraph{Hierarchical feature extractor}
Inspired by \cite{MVANet}, we increase the spatial feature density by utilizing multiple views, essentially splitting the input image into $1 \times 1$, $2 \times 2$, and $4 \times 4$ non-overlapping patches.
Each patch is then resized to a fixed resolution of \(518 \times 518\), and DINOv2~\cite{DINOv2}
The features are extracted independently for every patch.
After feature extraction, the features are rearranged back into their corresponding spatial
layouts and uniformly resized to \(3 \times C \times 148 \times 148\).
Finally, we aggregate the three hierarchical feature maps by calculating the channel-wise mean, producing a dense global feature representation of shape \(C \times 148 \times 148\).

% ----------------
% Losses in Detail
% ----------------
\subsection{Losses in Detail}
\paragraph{Converting 3D Hair growth direction to 2D hair orientation}
Predicting the 3D growth direction of hair is difficult and often ambiguous, and might require more care in future work.
For initial experiments, we therefore perform 2D hair orientation estimation instead of estimating the 3D hair growth direction.
We first project the 3D vector $\mathbf{v}=(v_x,v_y,v_z)^\top$ onto the image plane and compute its in-plane orientation $\theta=\operatorname{atan2}(v_y, v_x)$.
We then map this directed orientation $\theta$ to a 2D undirected orientation encoding $v' \in \mathbb{R}^2$ using an angle-doubling formulation:
\begin{align}
  \mathbf{v}' = \lVert \mathbf{v} \rVert_2 
  \begin{pmatrix}
  \cos(2\theta) \\
  \sin(2\theta)
  \end{pmatrix}
\end{align}
The angle doubling ensures that $\theta$ and $\theta+\pi$ are mapped to the same encoded direction, and the vector representation removes angular discontinuities, enabling stable learning with standard Euclidean losses.

\paragraph{Core loss functions}
We employ three core losses across all prediction tasks.  
Given a prediction $\hat{y} \in \mathbb{R}^{C \times H \times W}$, its corresponding ground-truth label $y \in \mathbb{R}^{C \times H \times W}$, and a set of valid pixels $M$, we define the scale- and shift-invariant mean squared error (SSIMSE) loss $\mathcal{L}_{\text{ssimse}}$ as
\begin{align}
  &\mathcal{L}_{\text{ssimse}}(\hat{y}, y, M, s, t) \nonumber \\
   &= \frac{1}{C \cdot |M|} \sum_{h, w \in M}
  \lVert s \hat{y}[:, h, w] + t - y[:, h, w] \rVert_2^2
\end{align}
Here, $s$ and $t$ are scalar scale and shift parameters that align the prediction to the ground truth.
The normalized multiscale gradient matching loss $\mathcal{L}_{\text{nmg}}$, inspired by~\cite{MiDaS}, is defined as
\begin{align}
  &\mathcal{L}_{\text{nmg}}(\hat{y}, y, M, s, t)  \nonumber \\
  &= \sum_{k=1}^{K} \frac{1}{C \cdot |M^{(k)}|} \sum_{h, w \in M^{(k)}} \nonumber \\
  &\;\;\;\;\;(\lVert s \nabla_x \hat{y}^{(k)}[:, h, w] - \nabla_x y^{(k)}[:, h, w] \rVert_2^2 \nonumber \\
  &\;\;\;\;\;+\lVert s \nabla_y \hat{y}^{(k)}[:, h, w] - \nabla_y y^{(k)}[:, h, w] \rVert_2^2)
\end{align}
Here, $(\hat{y}^{(k)}, y^{(k)}, M^{(k)})$ denote predictions, ground truth, and valid pixels at scale $k$.  
Following~\cite{MiDaS}, we set $K=4$, reducing the spatial resolution by a factor of two at each level.
Finally, for categorical prediction tasks, we include a masked cross-entropy loss $\mathcal{L}_{\text{ce}}$:
\begin{align}
  \mathcal{L}_{\text{ce}}(\hat{y}, y, M)
  &= - \frac{1}{|M|}
     \sum_{h,w \in M}
       \sum_{c=1}^{C}
         y[c,h,w] \,
         \log \hat{y}[c,h,w]
\end{align}
\textbf{Task specific losses}.
With these core losses defined, we specify the loss for the matting $\mathcal{L}_\text{matte}$, hair category estimation $\mathcal{L}_\text{cat}$, 2D hair orientation estimation $\mathcal{L}_\text{ori}$, and disparity / inverse-depth estimation $\mathcal{L}_\text{d}$ tasks as follows
\begin{align}
  \mathcal{L}_\text{matte} &= \frac{1}{2}\mathcal{L}_{\text{ssimse}}(\hat{y}_\text{matte}, y_\text{matte}, M_\text{all}, 1, 0) \nonumber \\
  &+ \frac{1}{2}\mathcal{L}_{\text{nmg}}(\hat{y}_\text{matte}, y_\text{matte}, M_\text{all}, 1, 0)\\
  \mathcal{L}_\text{cat} &= \mathcal{L}_{\text{ce}}(\hat{y}_\text{cat}, y_\text{cat}, M_\text{hair})\\
  \mathcal{L}_\text{ori} &= \frac{1}{2}\mathcal{L}_{\text{ssimse}}(\hat{y}_\text{ori}, y_\text{ori}, M_\text{hair}, 1, 0) \nonumber \\
  &+ \frac{1}{2}\mathcal{L}_{\text{nmg}}(\hat{y}_\text{ori}, y_\text{ori}, M_\text{hair}, 1, 0)\\
  \mathcal{L}_\text{d} &= \frac{1}{2}\mathcal{L}_{\text{ssimse}}(\hat{y}_\text{d}, y_\text{d}, M_\text{hair}, s_\text{d}, t_\text{d}) \nonumber \\
  &+ \frac{1}{2}\mathcal{L}_{\text{nmg}}(\hat{y}_\text{d}, y_\text{d}, M_\text{hair}, s_\text{d}, t_\text{d})\\
\end{align}
where $M_\text{all}$ contains all pixels and $M_\text{hair}$ only the hair pixels as valid.
We estimate $s_\text{d}$ and $t_\text{d}$ via the least-squares criterion as described in \cite{MiDaS}.
Finally, we combine the losses for the \modelname-Attributes submodule in $\mathcal{L}_\text{Attr}$.
\begin{equation}
  \mathcal{L}_\text{Attr} = \frac{1}{3} \left(\lambda_\text{cat} \mathcal{L}_\text{cat} + \lambda_\text{ori} \mathcal{L}_\text{ori} + \lambda_\text{d} \mathcal{L}_\text{d} \right)
\end{equation}
We set ($\lambda_\text{cat}=0.1$, $\lambda_\text{ori}=0.1$, $\lambda_\text{d}=0.1$), which scales the losses to be at a similar range as the matting loss $\mathcal{L}_\text{matte}$.

\begin{figure*}
  \centering{\includegraphics[width=\textwidth]{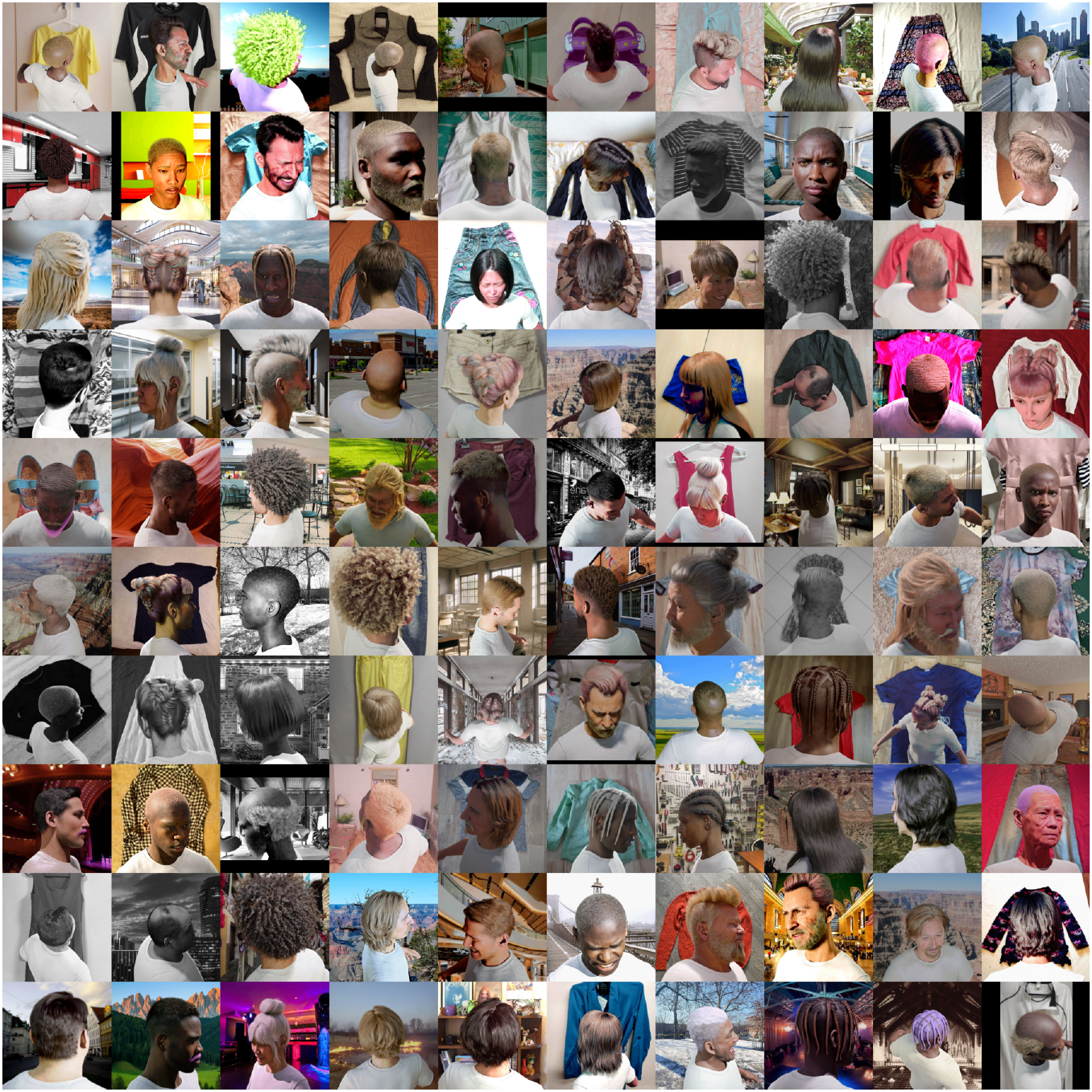}}
  \caption[Training images with augmentations]{\textbf{Training images with augmentations}. These figures show 100 random augmented samples from our \datasetname training set. We achieve a high variety in lighting and shading, despite using only one directional light source. For this illustration, we resize the image to $512 \times 512$ before we apply the augmentations to make resolution-dependent effects such as blur or noise more visible.}
  \label{fig:augmentation_examples}
\end{figure*}

\begin{figure*}
  \centering{\includegraphics[width=\textwidth]{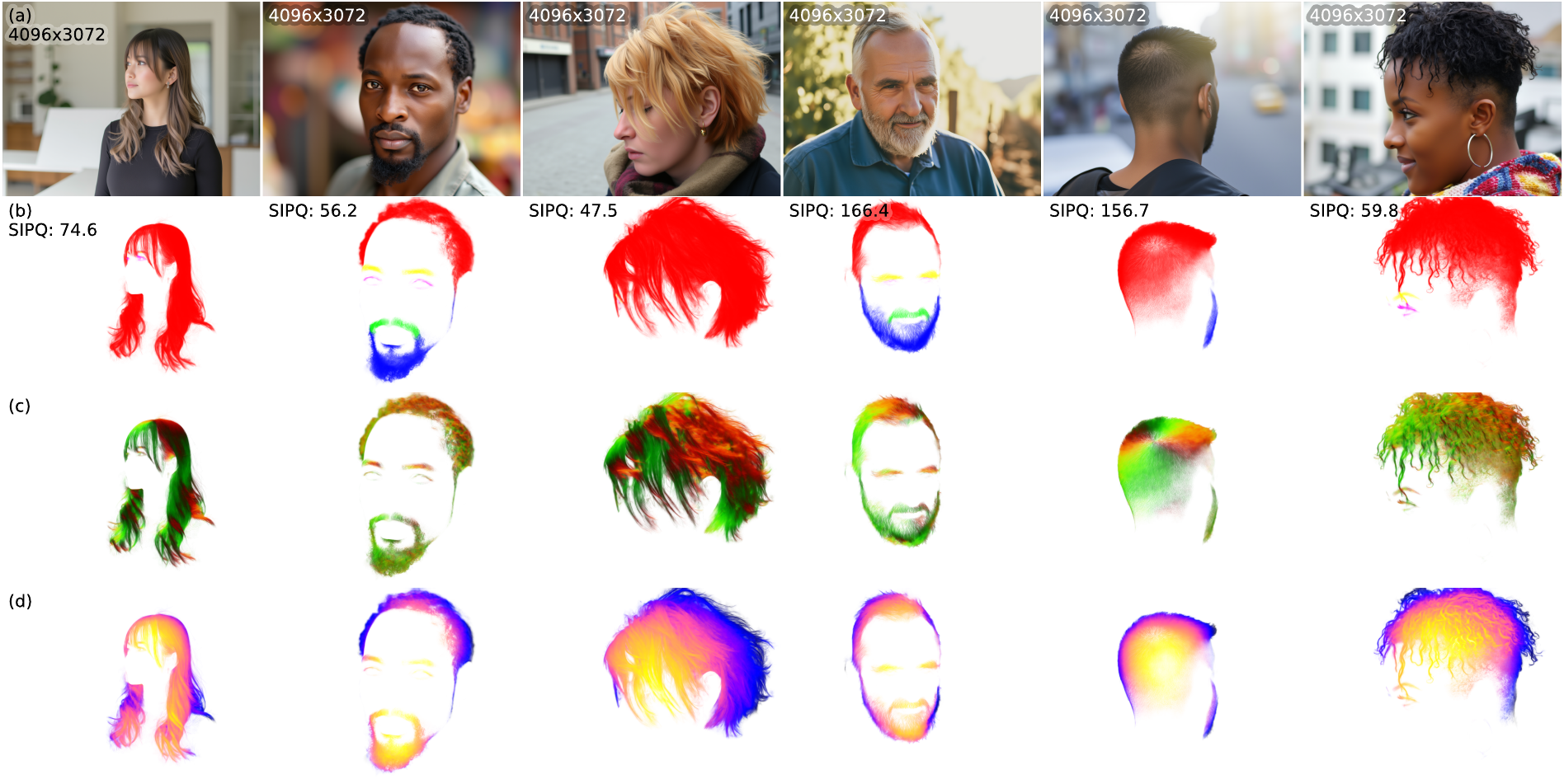}}
  \caption[Predictions of \datasetname's additional labels]{\textbf{High-resolution model predictions}. This figure shows additional predictions of both our \modelname-Matte and \modelname-Attributes models on images generated with Diffusion-4K~\cite{Diffusion4K}. (a) Input image. (b) Predicted hair matte, colored by the predicted hair category. (c) 2D hair orientation (d) Estimated relative depth.}
  \label{fig:model_predictions_examples}
\end{figure*}

% -------------
% Augmentations
% -------------
\subsection{Augmentations}
Given an RGBA image with labels, we first sample a random background image from one of two background datasets~\cite{BGMv2, ClothingDataset}, selecting between them with equal probability. We then perform histogram matching to adjust the lighting of the RGBA image to better match the sampled background before compositing it onto the background.
To increase visual diversity and improve robustness on real-world data, we randomly apply a set of augmentations, including \textbf{horizontal flipping}, \textbf{Gaussian noise}, \textbf{RGB-to-grayscale conversion}, \textbf{motion blur}, \textbf{defocus blur}, and \textbf{JPEG compression}, as well as random adjustments to the image \textbf{brightness}, \textbf{contrast}, \textbf{saturation}, \textbf{hue}, and \textbf{gamma}.
We show 100 random images with these augmentations in \cref{fig:augmentation_examples}.
Note that the lighting is consistent with the rendered MetaHuman as a result of the applied histogram-matching.

% -----------------
% Training schedule
% -----------------
\subsection{Training schedule}
Our models are trained on a single RTX 4090 GPU using the Adam optimizer with a batch size of 1.
Both modules of \modelname are trained using the same training schedule.
As illustrated in \cref{fig:model_architecture}, we keep the ViT backbone frozen and only train the \textit{Encoder} and \textit{Fusion} blocks.

\paragraph{Training on \datasetname}
For resolutions up to $2048 \times 2048$, we train using an initial learning rate of $1e^{-4}$ for the first 10 epochs and then reduce the learning rate to $1e^{-5}$ for the last 5 epochs.
For the 4K model version, we fine-tune the $2048 \times 2048$ \modelname at the full $4096 \times 4096$ image resolution using a learning rate of $1e^{-5}$ for 3 more epochs, as training at 4K resolution is significantly more time-consuming.
Training a full 4K model on all dataset labels with this 2-step approach takes about 4 to 5 days on a single RTX 4090 GPU.

\paragraph{Training on Real-World Datasets}
Figaro-1k~\cite{Figaro1k} and HiSa~\cite{HairStep} already provide predefined train and test splits.  
As CelebAMask-HQ~\cite{CelebAMask-HQ} does not include an official split, we create a custom split by randomly selecting 10k images for the test set.
The number of training epochs is chosen based on the highest mIoU achieved across all three test sets.
For Figaro-1k, HiSa, and CelebAMask-HQ, we train for 20, 20, and 7 epochs.

% --------------------------------------
% Qualitative Prediction of other labels
% --------------------------------------
\subsection{Qualitative Predictions of all labels}
\Cref{fig:model_predictions_examples} presents several qualitative examples across all five prediction tasks on synthetic ultra-high-resolution images generated with Diffusion-4K~\cite{Diffusion4K}.
Our \modelname effectively separates thin individual hair strands from the background, and we observe very fine details in the orientation and depth maps.

\begin{figure}[!ht]
  \centering
  \includegraphics[width=0.75\columnwidth]{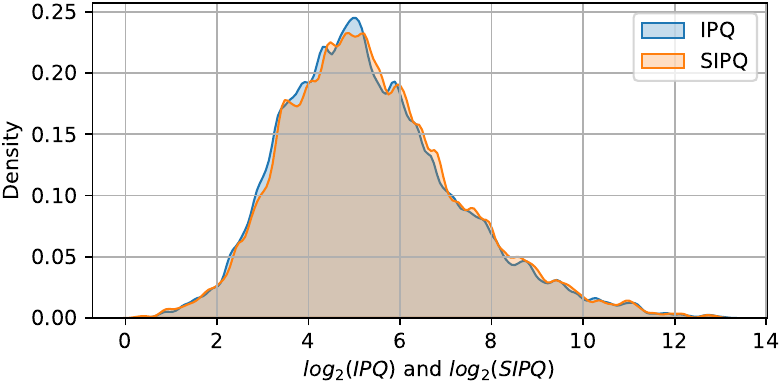}
  \caption{\textbf{IPQ vs. SIPQ}. Comparing the distribution of SIPQ vs. IPQ estimates on the binary segmentation dataset DIS5K~\cite{DIS5k}. The plot shows that the values of the generalized SIPQ match the original IPQ.}
  \label{fig:siqp_vs_ipq}
\end{figure}

\begin{figure}
  \centering{\includegraphics[width=\textwidth]{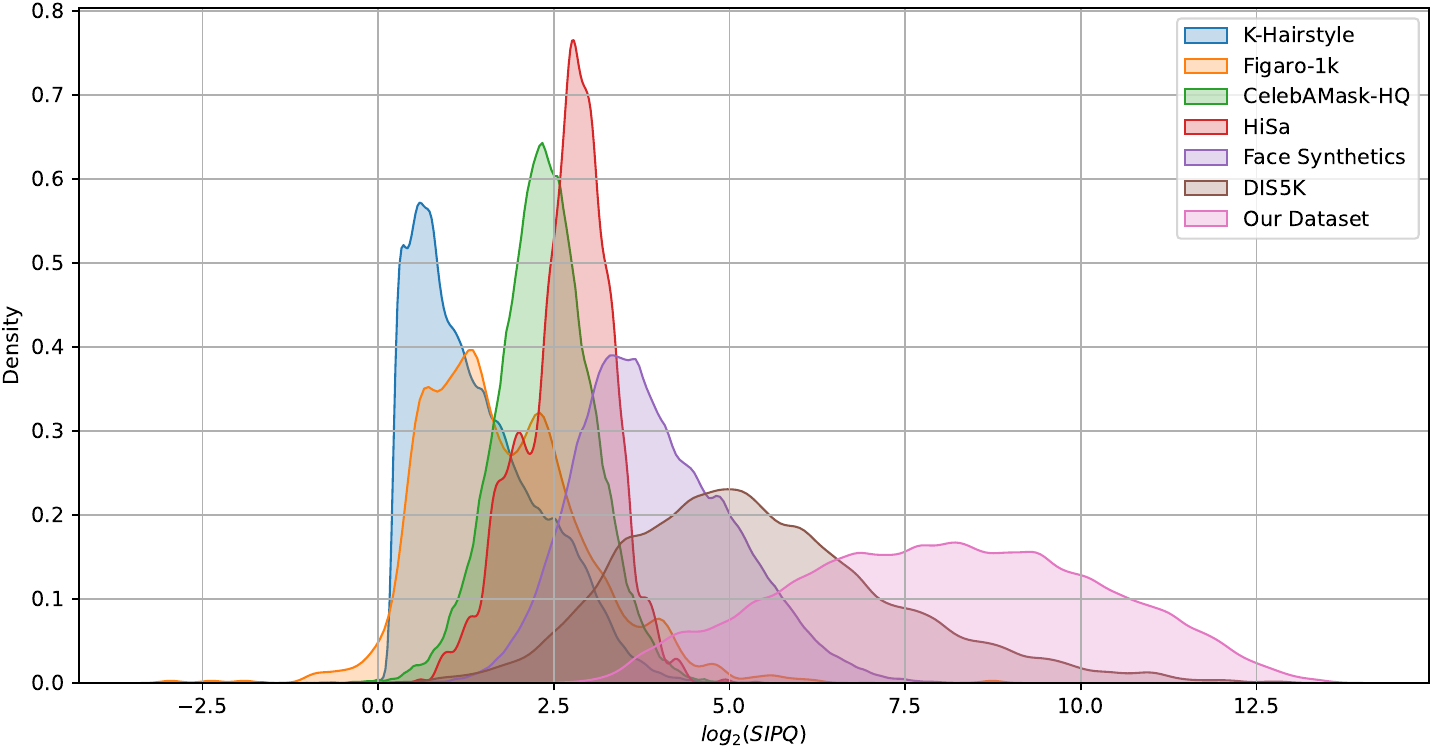}}
  \caption{\textbf{Datasets SIPQ Distributions}. Our \datasetname dataset results in samples with significantly higher SIPQ values. We plot the logarithm of the SIPQ values on the x-axis.}
  \label{fig:sipq_distributions}
\end{figure}

% -----------------------
% Region specific Matting
% -----------------------
\subsection{Region Specific Matting}
We measure the performance of \modelname, trained on \datasetname images, for different hair regions on real data. We evaluate across 300 randomly selected images and accumulate global statistics (TP/FP/FN across all images) to account for images without any eyebrows or beard hair, obtaining the results shown in \cref{tab:region_specific_generalization}. Lower scores are expected for thin eyebrows and beards with thin, short hair, which often cause noisy ground truth labels as shown in \cref{fig:datasets_matting_annotation_comparison}.

\begin{table*}[htbp]
    \caption{\textbf{Generalization on region specific evaluation}. We evaluate the domain generalization of \modelname trained on \datasetname in different hair categories. The mustache and beard masks are combined, as FaceSynthetics only provides one label for both categories.}
    \label{tab:region_specific_generalization}

    \centering
    \begin{tabular}{@{}lcccccc@{}}
        \toprule
        \multirow{2}{*}{Dataset} & \multicolumn{2}{c}{Scalp Hair} & \multicolumn{2}{c}{Eyebrows} & \multicolumn{2}{c}{Mustache $\cup$ Beard} \\
        \cmidrule(lr){2-3} \cmidrule(lr){4-5} \cmidrule(lr){6-7}
        & mIoU$\uparrow$ & F1$\uparrow$ & mIoU$\uparrow$ & F1$\uparrow$ & mIoU$\uparrow$ & F1$\uparrow$ \\
        \midrule
        CelebAMaskHQ~\cite{CelebAMask-HQ} & 89.36 & 94.38 & 66.20 & 79.66 & - & - \\
        FaceSynthetics~\cite{FaceSynthetics} & 88.47 & 93.88 & 46.35 & 63.34 & 62.42 & 76.86 \\
        \bottomrule
    \end{tabular}
\end{table*}

% -----------------------------------
% Sensitivity to number of MetaHumans
% -----------------------------------
\subsection{Ablation on Number of MetaHuman Identities}
We train a $512 \times 512$ version of H2M2 on varying numbers of MetaHuman identities to evaluate scaling behavior in \cref{tab:metahuman_identity_ablation}.
While our full training split currently contains 39 identities (43 in total including the validation split), doubling the count from 20 to 39 yields only marginal performance gains when evaluated on the Figaro1k dataset.
This saturation indicates that the number of MetaHuman identities in our dataset is no longer a primary performance bottleneck.

\begin{table*}[htbp]
    \centering
    \caption{\textbf{Number of MetaHuman Identities.} We evaluate the scalp hair matting performance on the Figaro1k dataset as the number of distinct MetaHuman models in the training set increases. The small improvement when increasing MetaHumans from 20 to 39 demonstrates that model performance effectively saturates.}
    \label{tab:metahuman_identity_ablation}
    \begin{tabular}{@{}lccc@{}}
        \toprule
        MetaHumans & mIoU$\uparrow$ & ACC$\uparrow$ & F1$\uparrow$ \\
        \midrule
        5 & 83.07 & 94.07 & 89.34 \\
        10 & 81.40 & 93.41 & 87.06 \\
        20 & 89.99 & 96.69 & 94.29 \\
        39 & 91.25 & 97.08 & 95.26 \\
        \bottomrule
    \end{tabular}
\end{table*}

% ---------------------
% Estimating Domain Gap
% ---------------------
\subsection{Estimating \datasetname Domain Gap}
We perform additional domain gap analysis based on 1000 samples per dataset.
For fairness, we resize the images to $512 \times 512$ resolution, use SAM~\cite{SAM} to isolate the person, then compute the feature vector with Inception-v3~\cite{InceptionV3}.
Masking out the background is important, as we are mainly interested in comparing the diversity and realism of the rendered MetaHumans and not the background.
To keep the comparison as fair as possible, for \datasetname we first compose the RGBA images on random backgrounds (without random augmentations) and then use the same feature-extraction process (including SAM for mask generation).

We compare datasets with both Fréchet Inception Distance~(FID)~\cite{FID} and Kernel Inception Distance~(KID)~\cite{KID} scores.
As shown in \cref{tab:realism_domain_gap}, SynthHair is most similar in feature distribution to Figaro1k (FID 112.1 / KID 80.8) and FaceSynthetics (FID 114.7 / KID 85.0).
This is similar to matching CelebAMaskHQ with Figaro1k (FID 119.6 / KID 87.5).

\begin{table*}[htbp]
    \caption{\textbf{Realism and Domain Gap}. We compare our \datasetname dataset with real portrait photographs on both FID and KID scores.}
    \label{tab:realism_domain_gap}

    \centering
    \resizebox{\linewidth}{!}{
        \begin{tabular}{@{}lcccccccccc@{}}
            \toprule
            \multirow{2}{*}{Dataset} & \multicolumn{2}{c}{SynthHair} & \multicolumn{2}{c}{Figaro1k} & \multicolumn{2}{c}{CelebAMaskHQ} & \multicolumn{2}{c}{HiSa} & \multicolumn{2}{c}{FaceSynthetics} \\
            \cmidrule(lr){2-3} \cmidrule(lr){4-5} \cmidrule(lr){6-7} \cmidrule(lr){8-9} \cmidrule(lr){10-11}
            & FID$\downarrow$ & KID$\downarrow$ & FID$\downarrow$ & KID$\downarrow$ & FID$\downarrow$ & KID$\downarrow$ & FID$\downarrow$ & KID$\downarrow$ & FID$\downarrow$ & KID$\downarrow$ \\
            \midrule
            SynthHair & - & - & 112.1 & 80.8 & 192.9 & 195.3 & 169.4 & 163.5 & 114.7 & 85.0 \\
            Figaro1k & 112.1 & 81.4 & - & - & 119.6 & 87.3 & 95.0 & 57.1 & 100.8 & 57.3 \\
            CelebAMaskHQ & 192.9 & 195.8 & 119.6 & 87.5 & - & - & 90.2 & 72.5 & 133.3 & 116.3 \\
            HiSa & 169.4 & 164.8 & 95.0 & 57.7 & 90.2 & 73.2 & - & - & 116.5 & 101.2 \\
            FaceSynthetics & 114.7 & 84.9 & 100.8 & 56.7 & 133.3 & 115.6 & 116.5 & 99.8 & - & - \\
            \bottomrule
        \end{tabular}
    }
\end{table*}

% ----------
% User Study
% ----------
\subsection{User Study Details}
\begin{figure}
    \begin{subfigure}[b]{1.0\linewidth}
        \caption{Side-by-side comparison}
        \includegraphics[width=\linewidth]{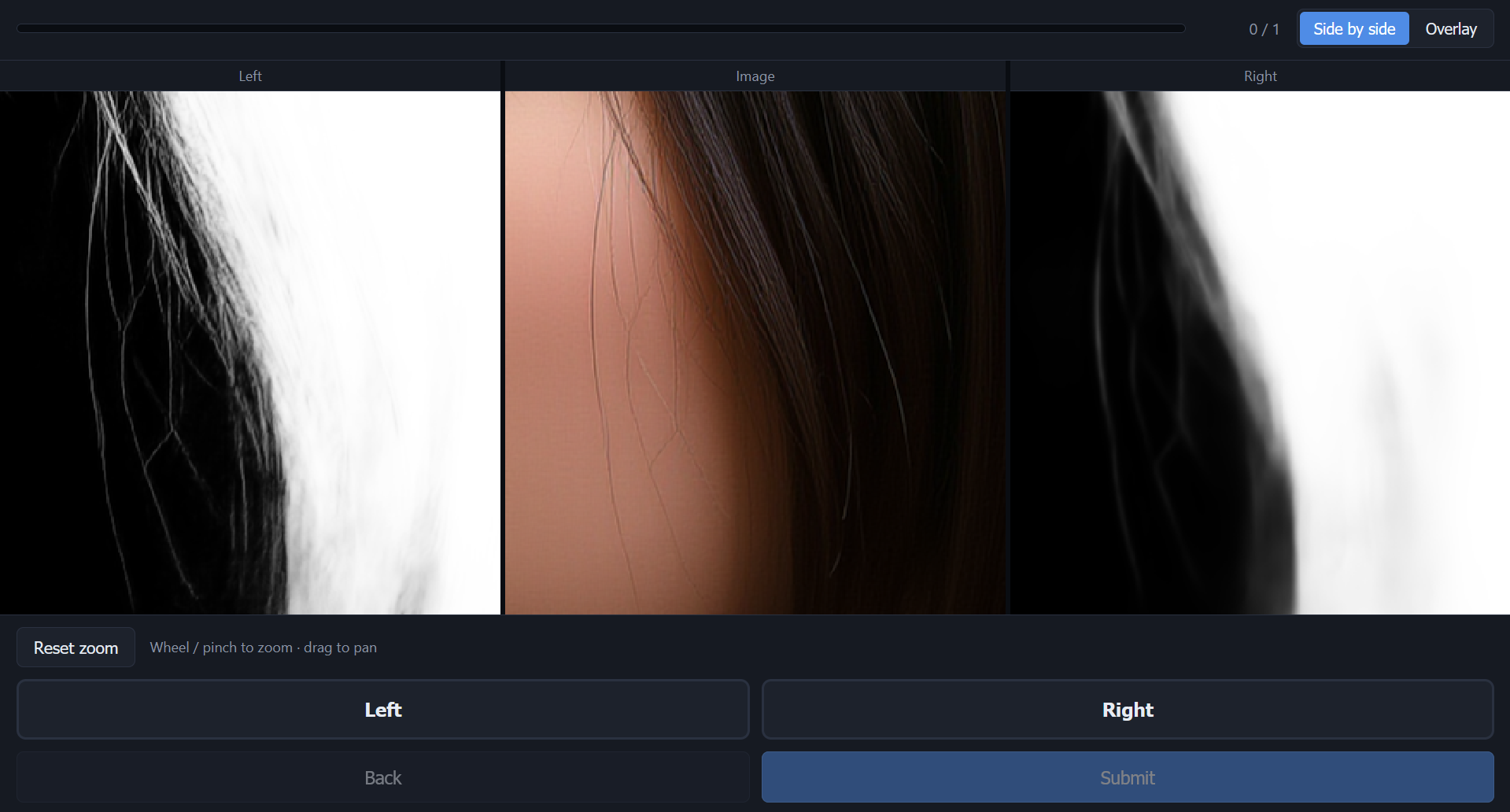}
    \end{subfigure}
    \begin{subfigure}[b]{1.0\linewidth}
        \caption{Overlay comparison}
        \includegraphics[width=\linewidth]{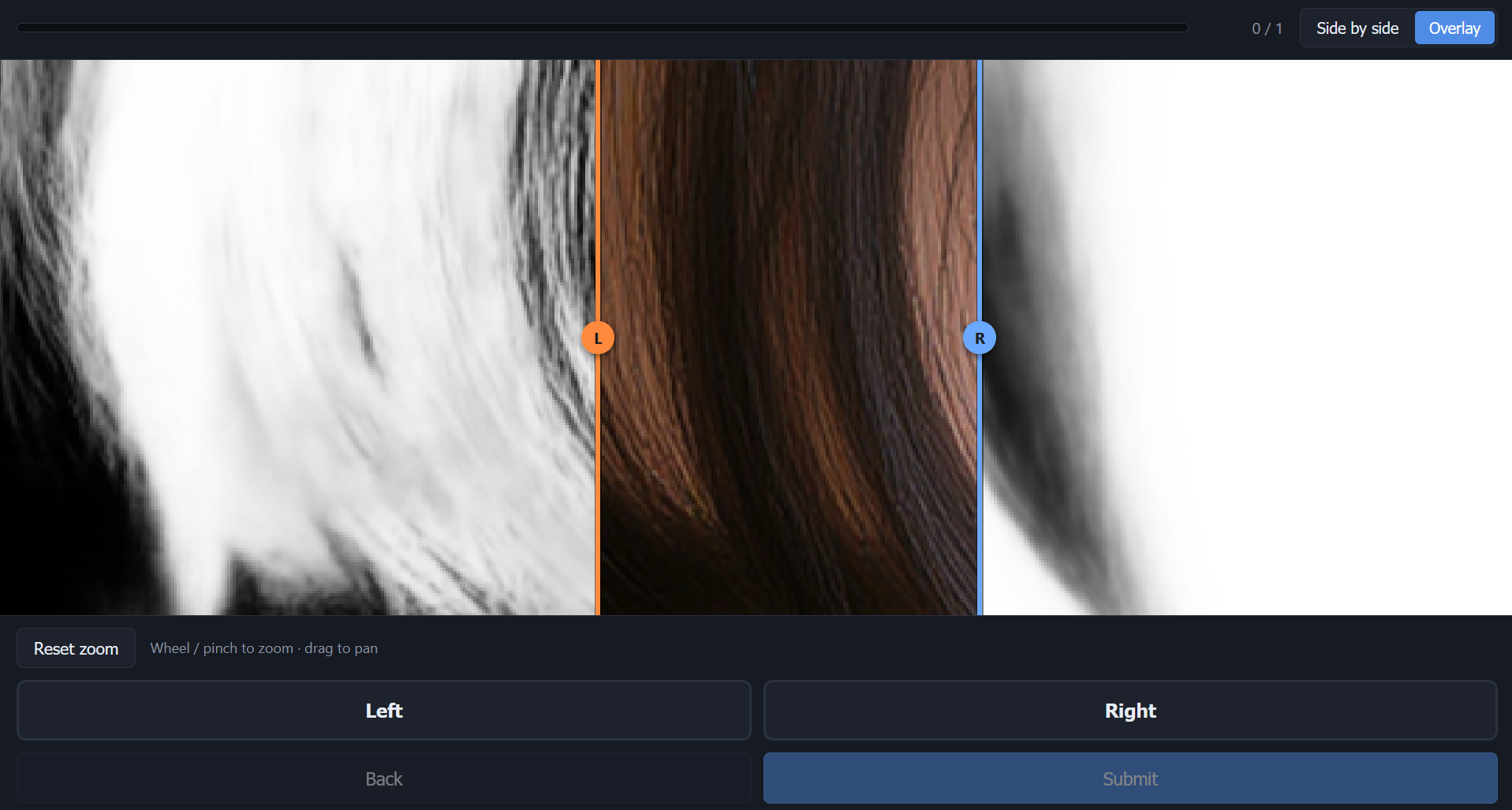}
    \end{subfigure}
    \caption{\textbf{A/B Tool}. Two zoomed-in screenshots of the interface used for our A/B user study. The left matte is from \modelname trained on \datasetname, while the right is from \modelname trained on CelebAMask-HQ. Users can choose between (a) a side-by-side and (b) an overlay view. Both views support zooming and panning to analyze matte details. The left and right sides are randomly shuffled for each comparison to prevent bias.}
    \label{fig:ab_tool}
\end{figure}

The user study is conducted with 15 participants on 30 randomly selected images of the HHM2K~\cite{SparseMat} dataset.
For each image, we create 4 predictions, resulting in a total of 90 A/B test cases.
The HHM2K dataset has a high set of diverse portrait images.
This results in a lot of images with little scalp hair to detect.
In order to select 30 good candidate images, we first sample 90 random images and compute, for each image, the mean of alpha density.
We then select the top 30 images with the highest hair density for our study.
In order to keep the participation time at about 10 minutes, we split the 90 test cases randomly into three groups, assigning 5 participants to each group.
During the test, each user is provided with the image and two possible hair matting options.
The user is instructed to "choose the option which better detects the scalp hair".
We show screenshots of the tool in \cref{fig:ab_tool}.
To estimate the statistical significance of our study, we compute exact binomial tests, analyzing both aggregate trials and per-user win rates.
Across all aggregate trials, our approach significantly outperforms \modelname trained on:
\begin{itemize}
    \item CelebAMask-HQ (98\% win rate): $p = 7.88 \times 10^{-40}$
    \item HiSa (87\% win rate): $p = 9.06 \times 10^{-22}$
    \item Figaro-1k (85\% win rate): $p = 2.22 \times 10^{-19}$
\end{itemize}
We also conduct binomial tests on the per-user average win rates.
All 15 evaluators independently favored our model over every baseline ($p = 6.10 \times 10^{-5}$ for each comparison).

% ----------------------
% Efficiency
% ----------------------
\subsection{Efficiency of \modelname}
To the best of our knowledge, neither \cite{TwoStageHumanHairSegmentation} nor \cite{FineGrainedHumanHairSegmentation} provides publicly available source code.
While \cite{TwoStageHumanHairSegmentation} reports an inference time of 15ms per image on the Figaro1k dataset (roughly $512 \times 512$ resolution) using an NVIDIA GTX 1080 Ti GPU, \cite{FineGrainedHumanHairSegmentation} does not report efficiency metrics.
Therefore, we benchmark \modelname directly against the hair matting baseline architecture PAN~\cite{PAN}.
Using the official GitHub implementation of PAN, we evaluate all models on an NVIDIA GeForce RTX 4090 GPU.
Experiments are run using randomly initialized tensors with a batch size of one and \texttt{bfloat16} precision. 
As detailed in \cref{tab:performance_resources}, the \modelname variants require significantly fewer parameters than PAN.
Furthermore, our approach scales much more efficiently to high resolutions (up to $4096 \times 4096$) in both runtime latency and peak VRAM allocation.

\begin{table*}[htbp]
    \caption{\textbf{Performance and VRAM usage}. We compare the total parameter count, latency in milliseconds per image, and peak VRAM (GiB) across different image resolutions.}
    \label{tab:performance_resources}
    \centering
    \begin{tabular}{@{}llcccccccc@{}}
        \toprule
        \multirow{2}{*}{Model} & \multirow{2}{*}{Parameters} & \multicolumn{4}{c}{Latency (ms / image)} & \multicolumn{4}{c}{Peak VRAM (GiB)} \\
        \cmidrule(lr){3-6} \cmidrule(lr){7-10}
        & & 512 & 1024 & 2048 & 4096 & 512 & 1024 & 2048 & 4096 \\
        \midrule
        H2M2-Matte & 22.94M & 24.4 & 27.5 & 48.8 & 147.1 & 0.32 & 0.33 & 0.50 & 1.79 \\
        H2M2-Attributes & 23.13M & 5.1 & 8.7 & 30.5 & 131.6 & 0.11 & 0.27 & 0.68 & 2.54 \\
        PAN & 168.96M & 6.9 & 21.0 & 82.0 & 333.3 & 0.44 & 0.55 & 1.25 & 4.04 \\
        \bottomrule
    \end{tabular}
\end{table*}

% -------------
% Cost Estimate
% -------------
\subsection{Cost Estimate}
\label{sec:estimated_cost}
Photorealistic rendering is computationally expensive.
The resolution of our dataset is significantly larger than that of Face Synthetics~\cite{FaceSynthetics}.
We estimate the cost per pixel, as we expect the computation cost to scale linearly with the number of pixels rendered. 
The SynthHairMan dataset~\cite{GenerativeVideoMatting} is most comparable to our work, as it is created with a similar resolution and rendered on the same RTX 4090 graphics card.\footnote{The dataset is actually generated in 3 different resolutions: $3840 \times 3840$, $3840 \times 2160$, and $2160 \times 3840$. Therefore, generating images at $4K \times 4K$ resolution might take longer than shown in \cref{tab:cost_estimate}.}
It uses Blender's Cycles to generate RGBA images with ray-tracing, and the resulting alpha channel is used as ground-truth human mask annotation.
The major difference to our work is that we do not only generate RGBA images, but also generate hair-matte, hair-category, hair growth, and depth maps, which require rendering the same scene multiple times.
To the best of our knowledge, the 3D head models of SynHairMan only contain scalp hair and do not have any skin or eye textures. 
This reduces the render complexity while significantly limiting the realism of images.

\begin{table*}[h!]
    \caption[Cost estimates of generating datasets]{\textbf{Cost Estimates of Generating Datasets.} Please note that despite the old age of the M60 GPU, we still observe a price of 1\$ per hour. Unfortunately, we could not find detailed information about the size of the SynHairMan dataset to provide a total cost estimate.}
    \label{tab:cost_estimate}
    \centering
    \begin{tabular}{@{}lllrrr@{}}
        \toprule
         \multirow{2}{*}{Dataset} & \multirow{2}{*}{Resolution} & \multirow{2}{*}{GPU} & Cost & Time & Total \\
          &  &  & per Hour & per Image & Cost\\
         \midrule
         FaceSynthetics~\cite{FaceSynthetics} & $512 \times 512$ & M60 & $\$1$ & $260$ sec.& $\$7$k\\
         SynthHuman~\cite{DAViD} & $512 \times 384$ & M60 & $\$1$ & $260$ sec. & $\$20$k\\
         SynHairMan~\cite{GenerativeVideoMatting} & $3840 \times 3840$ & RTX 4090 & $\$0.5$ & $60$ sec. & -\\
         \textbf{\datasetname (Ours)} & $4096 \times 4096$ & RTX 4090 & $\$0.5$ & $50$ sec. & $\$100$\\
         \bottomrule
    \end{tabular}
\end{table*}

% -------------------------
% Potential Societal Impact
% -------------------------
\subsection{Potential Societal Impact}
\label{sec:potential_societal_impact}
As with all human-centric vision tasks, our hair matting models may exhibit varying accuracy across demographic groups.
The use of synthetic data allows us to systematically address fairness concerns by controlling training distributions.
During the generation of our dataset, we randomized properties such as skin and hair texture to increase the diversity of our dataset.
Still, some forms of human diversity remain underrepresented, and undiscovered biases may persist.
Our synthetic dataset is generated efficiently with minimal resource requirements, reducing both computational and environmental costs without sacrificing robustness.
Additionally, we do not display any real images in our publication and instead use synthetically generated images with Diffusion-4K for our qualitative experiments.

%%%%%%%%%%%%%%%%%%%%%%%%%%%%%%%%%%%%%%%%%%%%%%%%%%%%%%%%%%%%

\newpage
\section*{NeurIPS Paper Checklist}

\begin{enumerate}

\item {\bf Claims}
    \item[] Question: Do the main claims made in the abstract and introduction accurately reflect the paper's contributions and scope?
    \item[] Answer: \answerYes{} % Replace by \answerYes{}, \answerNo{}, or \answerNA{}.
    \item[] Justification: Yes, the abstract and introduction both reflect the papers main contributions of analyzing existing hair matting and segmentation datasets, the \datasetname dataset, and experiments.

\item {\bf Limitations}
    \item[] Question: Does the paper discuss the limitations of the work performed by the authors?
    \item[] Answer: \answerYes{} % Replace by \answerYes{}, \answerNo{}, or \answerNA{}.
    \item[] Justification: The paper contains a limitations section detailing the limitations of SIPQ and our novel \datasetname dataset.

\item {\bf Theory assumptions and proofs}
    \item[] Question: For each theoretical result, does the paper provide the full set of assumptions and a complete (and correct) proof?
    \item[] Answer: \answerNA{} % Replace by \answerYes{}, \answerNo{}, or \answerNA{}.
    \item[] Justification: The paper focuses on dataset generation, empirical evaluation, and proposing a scale-invariant metric (SIPQ) rather than introducing new theoretical proofs.

    \item {\bf Experimental result reproducibility}
    \item[] Question: Does the paper fully disclose all the information needed to reproduce the main experimental results of the paper to the extent that it affects the main claims and/or conclusions of the paper (regardless of whether the code and data are provided or not)?
    \item[] Answer: \answerYes{} % Replace by \answerYes{}, \answerNo{}, or \answerNA{}.
    \item[] Justification: We provide the full source code for training the H2M2 models, running inference, and calculating the SIPQ metric in \url{https://github.com/mkarmann/synth-hair}. We also detail hyperparameters, architectural modifications, and data augmentation pipelines in the main text and appendix.

\item {\bf Open access to data and code}
    \item[] Question: Does the paper provide open access to the data and code, with sufficient instructions to faithfully reproduce the main experimental results, as described in supplemental material?
    \item[] Answer: \answerYes{} % Replace by \answerYes{}, \answerNo{}, or \answerNA{}.
    \item[] Justification: Code for training, inference, and metric calculation is provided in \url{https://github.com/mkarmann/synth-hair} as well as links for downloading the full dataset.

\item {\bf Experimental setting/details}
    \item[] Question: Does the paper specify all the training and test details (e.g., data splits, hyperparameters, how they were chosen, type of optimizer) necessary to understand the results?
    \item[] Answer: \answerYes{} % Replace by \answerYes{}, \answerNo{}, or \answerNA{}.
    \item[] Justification: Training schedules, hyperparameters, core loss functions, and data augmentation are explicitly outlined and explained in the appendix.

\item {\bf Experiment statistical significance}
    \item[] Question: Does the paper report error bars suitably and correctly defined or other appropriate information about the statistical significance of the experiments?
    \item[] Answer: \answerNo{} % Replace by \answerYes{}, \answerNo{}, or \answerNA{}.
    \item[] Justification: Training hair matting models in 4K is  computationally expensive. We are therefore not able to provide error bars for each experiment.

\item {\bf Experiments compute resources}
    \item[] Question: For each experiment, does the paper provide sufficient information on the computer resources (type of compute workers, memory, time of execution) needed to reproduce the experiments?
    \item[] Answer: \answerYes{} % Replace by \answerYes{}, \answerNo{}, or \answerNA{}.
    \item[] Justification: We specify the hardware (a single RTX 4090 GPU) and the time required for both dataset generation and model training.
    
\item {\bf Code of ethics}
    \item[] Question: Does the research conducted in the paper conform, in every respect, with the NeurIPS Code of Ethics \url{https://neurips.cc/public/EthicsGuidelines}?
    \item[] Answer: \answerYes{} % Replace by \answerYes{}, \answerNo{}, or \answerNA{}.
    \item[] Justification: The research conforms to the NeurIPS Code of Ethics as we utilize purely synthetic data generated via Unreal Engine for our dataset.

\item {\bf Broader impacts}
    \item[] Question: Does the paper discuss both potential positive societal impacts and negative societal impacts of the work performed?
    \item[] Answer: \answerYes{} % Replace by \answerYes{}, \answerNo{}, or \answerNA{}.
    \item[] Justification: We discuss the potential societal impacts, demographic representation, and the environmental/computational cost benefits of our synthetic data approach in the appendix.
    
\item {\bf Safeguards}
    \item[] Question: Does the paper describe safeguards that have been put in place for responsible release of data or models that have a high risk for misuse (e.g., pre-trained language models, image generators, or scraped datasets)?
    \item[] Answer: \answerNA{} % Replace by \answerYes{}, \answerNo{}, or \answerNA{}.
    \item[] Justification: The proposed dataset is purely synthetic and the models are specific to hair matting and hair attribute estimation.

\item {\bf Licenses for existing assets}
    \item[] Question: Are the creators or original owners of assets (e.g., code, data, models), used in the paper, properly credited and are the license and terms of use explicitly mentioned and properly respected?
    \item[] Answer: \answerYes{} % Replace by \answerYes{}, \answerNo{}, or \answerNA{}.
    \item[] Justification: We reached out to Epic Games to prevent licensing issues with the MetaHuman Creator and the Unreal Engine.

\item {\bf New assets}
    \item[] Question: Are new assets introduced in the paper well documented and is the documentation provided alongside the assets?
    \item[] Answer: \answerYes{} % Replace by \answerYes{}, \answerNo{}, or \answerNA{}.
    \item[] Justification: We introduce the SynthHair dataset. Detailed documentation regarding the dataset's composition, the types of ground truth labels provided, and the exact pipeline used to generate it is also covered in the appendix.

\item {\bf Crowdsourcing and research with human subjects}
    \item[] Question: For crowdsourcing experiments and research with human subjects, does the paper include the full text of instructions given to participants and screenshots, if applicable, as well as details about compensation (if any)? 
    \item[] Answer: \answerYes{} % Replace by \answerYes{}, \answerNo{}, or \answerNA{}.
    \item[] Justification: The A/B user study was conducted with 15 unpaid volunteers (peers and colleagues) who freely chose to participate. The study was brief (approximately 10 minutes). Because no crowdsourcing platforms or hired labor were utilized, the minimum wage guideline for data workers does not apply here. Details on the instructions given to the volunteer participants are included in the Appendix.

\item {\bf Institutional review board (IRB) approvals or equivalent for research with human subjects}
    \item[] Question: Does the paper describe potential risks incurred by study participants, whether such risks were disclosed to the subjects, and whether Institutional Review Board (IRB) approvals (or an equivalent approval/review based on the requirements of your country or institution) were obtained?
    \item[] Answer: \answerNA{} % Replace by \answerYes{}, \answerNo{}, or \answerNA{}.
    \item[] Justification: The user study consisted entirely of an A/B visual preference evaluation of generated images. No personally identifiable information was collected, and the study posed no risk to participants, which typically exempts it from formal IRB review.

\item {\bf Declaration of LLM usage}
    \item[] Question: Does the paper describe the usage of LLMs if it is an important, original, or non-standard component of the core methods in this research? Note that if the LLM is used only for writing, editing, or formatting purposes and does \emph{not} impact the core methodology, scientific rigor, or originality of the research, declaration is not required.
    %this research? 
    \item[] Answer: \answerNA{} % Replace by \answerYes{}, \answerNo{}, or \answerNA{}.
    \item[] Justification: No Large Language Models were used as a core, original, or non-standard component of the methodology presented in this research.

\end{enumerate}

\end{document}